\documentclass[onecolumn]{arxiv}
\usepackage{lineno}
\usepackage{xcolor,colortbl}
\modulolinenumbers[5]
\usepackage{yaacro}
\usepackage{subfigure}
\usepackage{amsfonts} 
\usepackage{amsmath}
\usepackage{url}
\usepackage[colorinlistoftodos]{todonotes}
\presetkeys{todonotes}{inline,backgroundcolor=yellow}{}

\begin{acgroupdef}[list=acronimos]
    \acdef{ANN}{Artificial Neural Network}
    \acdef{A3C}{Asynchronous Advantage Actor Critic}
    \acdef{LiDAR}{Light Detection And Ranging}
    \acdef{UAV}{Unmanned Aerial Vehicle}
    \acdef{RL}{Reinforcement Learning}
    \acdef{DRL}{Deep Reinforcement Learning}
    \acdef{CNN}{Convolutional Neural Network}
    \acdef{DDPG}{Deep-Deterministic Policy Gradient}
    \acdef{TD3}{Twin Delayed Deep-Deterministic Policy Gradient}
    \acdef{DDQN}{Double Deep $Q$-Network}
    \acdef{DQN}{Deep $Q$ Network}
    \acdef{DPPO}{Distributed Proximal Policy Optimization}
    \acdef{SAC}{Soft Actor-Critic}
    \acdef{ReLU}{Rectified Linear Unit}
    \acdef{ROS}{Robot Operating System}
	\acdef{CNPq}{Conselho Nacional de Desenvolvimento Científico e Tecnológico}
	\acdef{CAPES}{Coordenação de Aperfeiçoamento de Pessoal de Nível Superior}
	\acdef{FAPEMIG}{Fundação de Amparo à Pesquisa do Estado de Minas Gerais}
\end{acgroupdef}

\begin{document}

%Space above and below the equations
\setlength{\abovedisplayskip}{5pt}
\setlength{\belowdisplayskip}{5pt}

%\title{On the Generalization of Deep Reinforcement Learning Methods in the Problem of Local Navigation}
\title{Generalization in Deep Reinforcement Learning for Robotic Navigation by Reward Shaping}

\author{
Victor R. F. Miranda$^{1}$
\and
Armando A. Neto$^{2}$
\and
Gustavo M. Freitas$^{3}$
\and
Leonardo A. Mozelli$^{2}$
}

\institute{$^{1}$ 
              Graduate Program in Electrical Engineering (PPGEE), Universidade Federal de Minas Gerais (UFMG), Belo Horizonte, MG, Brazil. \email\texttt{victormrfm@ufmg.br}\\
           $^{2}$ 
              Department of Electronics Engineering (DELT), UFMG, Belo Horizonte, MG, Brazil. \email{\texttt{\{aaneto,mozelli\}@cpdee.ufmg.br}}\\
            $^{3}$
              Department of Electrical Engineering (DEE), UFMG, Belo Horizonte, MG, Brazil. \email{\texttt{gustavomfreitas@ufmg.br}}
              }

\date{}

\color{black}

\maketitle

\begin{abstract}
    In this paper, we study the application of \ace{DRL} algorithms in the context of local navigation problems, in which a robot moves towards a goal location in unknown and cluttered workspaces equipped only with limited-range exteroceptive sensors, such as \ac{LiDAR}. Collision avoidance policies based on \ac{DRL} present some advantages, but they are quite susceptible to local minima, once their capacity to learn suitable actions is limited to the sensor range. Since most robots perform tasks in unstructured environments, it is of great interest to seek generalized local navigation policies capable of avoiding local minima, especially in untrained scenarios.
    To do so, we propose a novel reward function that incorporates \emph{map information} gained in the training stage, increasing the agent's capacity to deliberate about the best course of action. Also, we use the \ace{SAC} algorithm for training our \ace{ANN}, which shows to be more effective than others in the state-of-the-art literature. A set of \emph{sim-to-sim} and \emph{sim-to-real} experiments illustrate that our proposed reward combined with the \ac{SAC} outperforms the compared methods in terms of local minima and collision avoidance.

\end{abstract}

%%%%%%%%%%%%%%%%%%%%%%%%%
% operacoes
\newcommand{\escalar}[1]{\ensuremath{\mathit{#1}}}
\newcommand{\vetor}[1]{\ensuremath{\boldsymbol{#1}}}
\newcommand{\matriz}[1]{\ensuremath{\mathbf{\uppercase{#1}}}}
\newcommand{\conjunto}[1]{\ensuremath{\mathcal{\uppercase{#1}}}}
\newcommand{\distribuicao}[1]{\ensuremath{\mathcal{\uppercase{#1}}}}
\newcommand{\transpose}{\ensuremath{{}^\intercal}}

%%%%%%%%%%%%%%%%%%%%%%%%%
\newcommand{\statev}{\vetor{s}}
\newcommand{\laser}{\vetor{l}}
\newcommand{\pos}{\vetor{p}}
\newcommand{\vel}{\vetor{v}}

\newcommand{\actionv}{\vetor{a}}

\section{Introduction}
\label{sec:intro}

Among the current problems in Mobile Robotics, \emph{safe navigation} and \emph{exploration} of unknown scenarios are two of the most challenging ones. In particular, exploration missions, like search and rescue, precision farming, or caving, are very important in military and civilian field robotics applications. But, although there is a vast literature on those subjects, several questions still demand better solutions, such as planning on cluttered unknown spaces and real-time trajectory optimization. 

In the last few years, with the recent advent of deep neural networks, \ace{DRL} algorithms have been widely employed in several problems of autonomous systems \cite{Mnih2015Humanlevel}. But despite the rapid progress, endowing robots with the capability of learning skills in such complex and unstructured scenarios remains challenging, as it is quite difficult to train \emph{model-free} policies in growing spaces with a large set of possible obstacle configurations. To be effective, \ace{DRL} controllers must be potentially generalized. In other words, they must be capable of realizing missions (with relative success) not only in those situations in which the police has be trained but also in untrained contexts, avoiding the phenomenon of overfitting.

Several works address the generalization problem in \ac{DRL}~\cite{kirk2021survey}, proposing different solutions such as: increasing the similarity between training and execution environments with data augmentation, domain randomization, or generating different environments along training; and handling differences between environments with traditional regularization techniques. However, these generalization strategies require wide data variation and training steps to cover all characteristics of different environments. Still, the algorithms can find problems when running in a different environment from those used in training (simulation to the real world, for example).

Here, our proposed neural network is trained using the reinforcement learning method with the \ace{SAC} algorithm and a collision avoidance navigation policy. \ac{SAC} is an off-policy actor-critic algorithm for continuous actions that optimizes a stochastic policy. Despite being derived from \ace{DDPG} \cite{lillicrap2015continuous} and \ac{TD3}, \ac{SAC} seeks to maximize the policy entropy together with the reward. This strategy increases the exploration of actions (vary the actions to find useful learning) during training, helping to deal with local minima problems in cluttered environments and to adapt to untrained situations.

Furthermore, still seeking to reduce local minima problems, we propose a novel reward function that penalizes actions that don't increase information in a local map exploration and don't reduce the distance to the goal location. Therefore, we expect that the agent does not receive a reward when selecting actions that maintain the robot stuck or moving around the same place, which characterizes a local minimum. Finally, our reward function also penalizes collisions and congrats the agent for arriving at the target, moving in clear directions (with no obstacles or away from them), and performing progressive movements (not standing still). 

The observation state of the proposed training strategy uses data from \ac{LiDAR} and distance to the target point. The local map is only used in the reward function, which does not affect the navigation system performance during the execution stage.

When compared to the current literature, we aim for the following contributions:
\begin{itemize}

    \item a novel reward function that incorporates exteroceptive information (newer perception information from the environment map) and improves the agent's robustness to the local minimums of the environment;
    
        \item a comparative analysis between the model trained with the proposed novel reward function and others from the literature;
    
    \item a comparative analysis between two \ac{DRL} algorithms, the \ace{TD3}~\cite{fujimoto2018addressing} and the \ac{SAC}~\cite{haarnoja2018soft}, in which we demonstrate that the \ac{SAC} outperforms the \ac{TD3} in the context of safe local navigation for mobile robots with continuous actions;
    
    \item a set of \emph{sim-to-sim} and \emph{sim-to-real} experiments to evaluate the generalization of our proposed approach.
\end{itemize}

The remainder of this paper is organized as follows. Section~\ref{sec:related} presents the related work, especially focusing on \ac{DRL} navigation and exploration methods. Section~\ref{sec:methodology} discusses the problem formalization, presents the proposed solution, and details the formal analysis. Simulated results are shown in Section~\ref{sec:results}, while the final remarks are presented in Sec.~\ref{sec:conclusion}.
\section{Related work}
\label{sec:related}

In the last decades, autonomous navigation tasks have been largely studied in the context of Mobile Robotics. Several approaches, ranging from reactive to deliberative hierarchies, have been used to endow the robots with the capacity to explore and modify a given scenario.
%
%Mobile robots have been adopted as a solution for several automation tasks, increasing the interest in autonomous navigation systems. 
Many researchers propose control methods to move a robot from its location to a target point in the environment, avoiding obstacles until reaching the objective. Classical strategies are based on following a plan minimizing the position or the distance error between the robot and the desired path or trajectory \cite{sanchez2021nonlinear,Rezende2020,Pereira2021,ZHANG2020103565}. However, these methods require a previously planned path/trajectory to be followed, which may demand prior knowledge of the environment map and obstacle location to compute a clear path.

Other control strategies do not use prior planned paths but define actions to navigate towards the target, avoiding locally detected obstacles. The authors of \cite{Sivaranjani2021} propose an investigation of bug algorithms, comparing them in simulated environments. These algorithms define control actions to navigate toward the goal avoiding obstacles that obstruct the robot's movements. Similarly, the authors in \cite{Sfeir2011,SUDHAKARA2018998} present a method using potential fields to compute velocities for the robot to navigate in the goal direction without colliding with obstacles. However, these strategies can present problems, such as local minimum, when performing in complex scenarios.

Considering the advances in Machine Learning, Deep Neural Networks, and Reinforcement Learning, researchers began to investigate the use of \ac{DRL} as a basis for control~\cite{KASAEI2021103900,CARLUCHO201871}, navigation~\cite{PATEL2019,Kahn2018}, localization~\cite{Gao2020,Bohez2017}, and planning~\cite{reinhart2020learning,YOU20191} systems.
The versatility of neural networks, associated with the possibility of training based only on metrics and heuristics as rewards for actions taken, attracts current research to adopt these techniques in Robotics. 

Regarding autonomous robot navigation, \ac{DRL} surveys address the efficiency of models trained using reinforcement learning and the various algorithms developed for this application \cite{Zhu2021,Jiang2020}.

In \cite{Ruan2019}, for example, the authors use discrete prior defined actions as options to train a policy model with a \ac{DDQN} algorithm, considering only camera images as input to navigate through an environment while avoiding obstacles. Although it works in many scenarios, discrete actions can be a problem in environments with a high number of obstacles. Besides that, the reward function proposed by the authors does not consider local minimum problems. Similarly, \cite{chen2021deep} presents a navigation strategy also using a policy trained with the \ac{DDQN} algorithm and discrete actions. The reward function also depends only on the distance between the robot and target, grants for arrival, and penalizes collisions. Additionally, the proposed method requires an occupancy grid map as input, increasing the computational processing on runtime. 

Still using discrete actions, the authors of \cite{Liu2020} propose a navigation policy trained using the \ac{A3C} that allows training a global agent from parallel agents in learning. In terms of generalization, this approach helps to cover a large number of maps or navigation situations, despite the discrete actions. Our proposed strategy, on the other hand, focuses on avoiding static and dynamic obstacles by utilizing data from an occupancy map and \ac{LiDAR}, which increases computation processing for training and runtime.

Considering continuous actions of velocities for the robot navigation, the authors in \cite{grando2021deep} present a method that uses the \ac{DDPG} algorithm to train a policy to control a hybrid \ac{UAV} to navigate towards a target point, avoiding local detected obstacles. The model uses the robot's states, \ac{LiDAR} measurements, and distance to the target point, to select the best control actions for the \ac{UAV}. Despite the qualitative results presented, the chosen reward function is too simple and may result in poor performance in more complex scenarios.

In the map exploration context, \cite{cimurs2021goal} and \cite{hu2020voronoi} present strategies to explore the space using a navigation system with a trained policy. The first one uses the \ac{TD3} algorithm to train and offers a reward function that observes only arrivals, collisions, and nonprogressive movement conditions. For the second one, the authors use the \ac{DDPG} for training and a reward that includes a safety clearance term, avoiding obstacles, and moving in the target direction. Since the robot must navigate the environment across long distances, local minimum issues are frequently seen in map exploration tasks. The authors of \cite{cimurs2021goal} propose auxiliary algorithms to help in the target destination to avoid local minimum and increase the method generalization, and \cite{hu2020voronoi} use a safety clearance reward to prevent a high attraction of the robot to the target. However, the algorithms used for training do not have high action exploration during the agent learning, and the proposed reward function does not deal with this issue directly.

Generalization is a paradigm for neural network model training. In the context of reinforcement learning, several studies discuss the capability of models in actuating on untrained environments or situations and address methods to improve their efficiency \cite{kirk2021survey}.
Some papers present forms of quantifying, measuring, and characterizing the generalization in \ac{DRL} methods \cite{cobbe2019quantifying,witty2021measuring}, and others focus on developing strategies to increase the generalization capability of the models by modifying the learning algorithms, as in the present paper. Another concern in terms of generalization is the transfer of simulation-trained models to real-world situations. Similarly to \cite{tai2017virtual}, we also address the sim-to-real problem. 

The model generalization is also linked to its exploration capability, i.e., vary the actions during training to find useful learning. Exploration is generally a challenge in Reinforcement Learning, but an important technique for sparse problem applications. It is common in most advanced learning algorithms to include techniques for exploration, some more advanced than others. There are different techniques to increase exploration during training, as presented in \cite{LADOSZ20221}.  

% Survey Generalization - \cite{kirk2021survey}

% \cite{tai2017virtual} - Virtual to Real, DDPG, recompensa - distancia, chegou, bateu

%\section{Methodology}
\section{Deep Reinforcement Learning setup}
\label{sec:methodology}

The navigation problem generally consists of computing a sequence of commands that moves the robot along a path represented by a curve or a sequence of waypoints in the workspace. Furthermore, we can incorporate safe navigation to this problem when facing environments with obstacles and then using obstacle-avoidance techniques. The most common strategies for safe navigation involve planning and motion control steps, which often require the environment map and impose a high computational cost. In our approach, a trained neural network commands the robot to navigate the unknown map avoiding obstacles until reaching the target position.

\begin{figure}[htpb]
    \centering
     \includegraphics[trim={0cm 4.0cm 0cm 0.8cm},clip,width=1\textwidth]{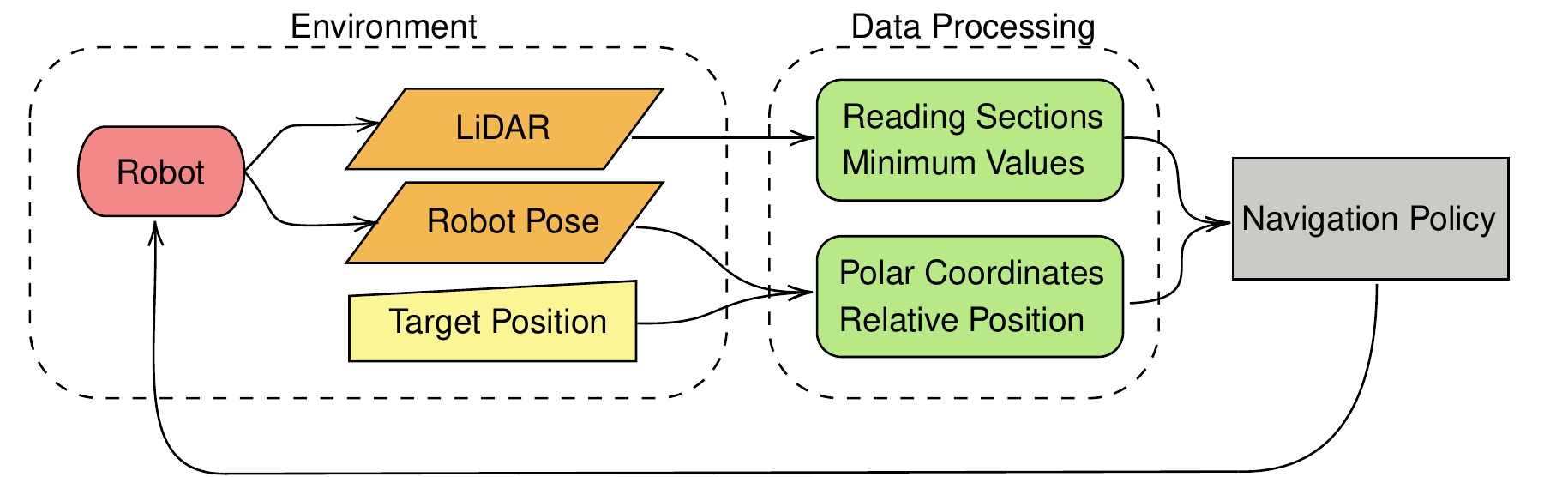}
    \caption{Proposed navigation system overview.}
    \label{fig:system_diagram}
\end{figure}

Fig.~\ref{fig:system_diagram} illustrates the complete procedure of the proposed navigation system. First, sensors collect data from the robot's states and the environment, and also a goal point is defined. In the sequence, the system processes these data in order to create the observation state for the trained navigation policy model. Finally, the policy defines actions of velocities for the robot based on the states observed.

%%%%%%%%%%%%%%%%%%%%%%%%%%%%%%%%%%%%%%%%%%%%%%%%%%
%%%%%%%%%%%%%%%%%%%%%%%%%%%%%%%%%%%%%%%%%%%%%%%%%%
%\subsection{Deep Reinforcement Learning Setup}
% The proposed neural network is trained using the reinforcement learning method with the \ac{SAC} algorithm and a collision avoidance navigation policy. \ac{SAC} is an off-policy actor-critic algorithm for continuous actions that optimizes a stochastic policy. Despite being derived from \ace{DDPG}\cite{lillicrap2015continuous} and \ac{TD3}, \ac{SAC} seeks to maximize the entropy of the policy together with the reward. This strategy increases the exploration of actions during training, helping to deal with local minimums problems in robot navigation environments.

%%%%%%%%%%%%%%%%%%%%%%%%%%%%%%%%%%%%%%%%%%%%%%%%%%
%%%%%%%%%%%%%%%%%%%%%%%%%%%%%%%%%%%%%%%%%%%%%%%%%%
%\subsubsection{Observation States and Actions}
\subsection{Observation state and action representation}
\label{subsec:obs_statesDescription}
The observation state vector $\statev$ includes the robot's relative position to the target region $\pos \in \mathbb{R}^2$, absolute linear $v$ and angular $\omega$ velocities, and the information data provided by the robot's planar LiDAR $\laser \in \mathbb{R}^l$, with $l$ being the number of laser beams. Formally, it can be represented as:
\begin{equation}
    %\statev_{t} = \left[ \laser_{t},\ \pos_{t},\ \vel_{t-1} \right],
    \statev = 
    \begin{bmatrix} 
        \pos\\ v\\ \omega\\ \laser
    \end{bmatrix}.
\end{equation}

On the other hand, the continuous action vector includes the commanded linear velocity $\bar{v} \in \mathbb{R}_{+}$ and commanded angular velocity $\bar{\omega} \in \mathbb{R}$, such that it can be represented by:
\begin{equation}
    \actionv = 
    \begin{bmatrix} 
        \bar{v} \\ \bar{\omega}
    \end{bmatrix}.
\end{equation}

%%%%%%%%%%%%%%%%%%%%%%%%%%%%%%%%%%%%%%%%%%%%%%%%%%
%%%%%%%%%%%%%%%%%%%%%%%%%%%%%%%%%%%%%%%%%%%%%%%%%%
%\subsubsection{Reward}
\subsection{Reward function}

An important part of reinforcement learning concerns defining a reward function for the policy. This function will guide the agent to select correct actions according to the observation state. Therefore, the reward must be well defined according to the learning objective in the environment.

We propose a reward function for collision avoidance navigation policy that encourages the robot to take actions that will move it towards the target together while keeping away from local minimums of the environment. Combining environmental information extracted via sensors with exteroceptive information provided by auxiliary algorithms, the proposed reward merges four different terms in a system described by:
\begin{equation}
r = 
\left\{
\begin{aligned}
& r_a & \mathrm{if} \ d \leq d_{min}, \\
& r_c & \mathrm{if \ collision}, \\
& r_t & \mathrm{if \ steps} \geq \mathrm{Timeout}, \\
& r_{Gd} + r_{l} + r_{v} & \mathrm{otherwise},
\end{aligned}
\right.
\end{equation}
where $r_a$ is a positive reward given when the Euclidean distance $d$ between the robot and the target is less than or equal to a minimal value $d_{min}$, $r_c$ is a penalty reward given in case of collision, and $r_t$ is a negative reward given if the number o steps exceed a \emph{Timeout} limit. 
In other cases, the reward adds three different values respecting the following equations:
\begin{align}
    & r_{Gd} = \frac{G}{d},\\
    & r_l = \mathrm{min}(lidar|_{l_1}^{l_2}), \label{eq:r_l}\\
    & r_v = v - |\omega|,
\end{align}
where $r_{Gd}$ represents the ratio between the increased map information $G$ since the last measurement (with regards to newly mapped areas) and $d$, $r_l$ is the minimum value measured by the LiDAR in an interval between $l_1$ and $l_2$ representing a read in the robot's front side, and $r_v$ is a reward based on the linear and angular velocities.

The first part $r_{Gd}$ prevents or reduces local minimum problems, providing a reward only if the robot is moving to new places (avoiding already visited places that do not allow approaching the destination point) and increasing this reward value as the distance between the robot and the target decreases. To encourage actions moving into free spaces, $r_l$ gives a reward according to the minimum LiDAR measure in front of the robot. Finally, aiming to prevent movements without progression, $r_v$ provides a reward following the robot's velocities \cite{cimurs2021goal}.

%%%%%%%%%%%%%%%%%%%%%%%%%%%%%%%%%%%%%%%%%%%%%%%%%%
%%%%%%%%%%%%%%%%%%%%%%%%%%%%%%%%%%%%%%%%%%%%%%%%%%
%\subsubsection{Network Structure}
\subsubsection{Network structure}

\begin{figure}[t]
    \centering
     \includegraphics[width=\linewidth]{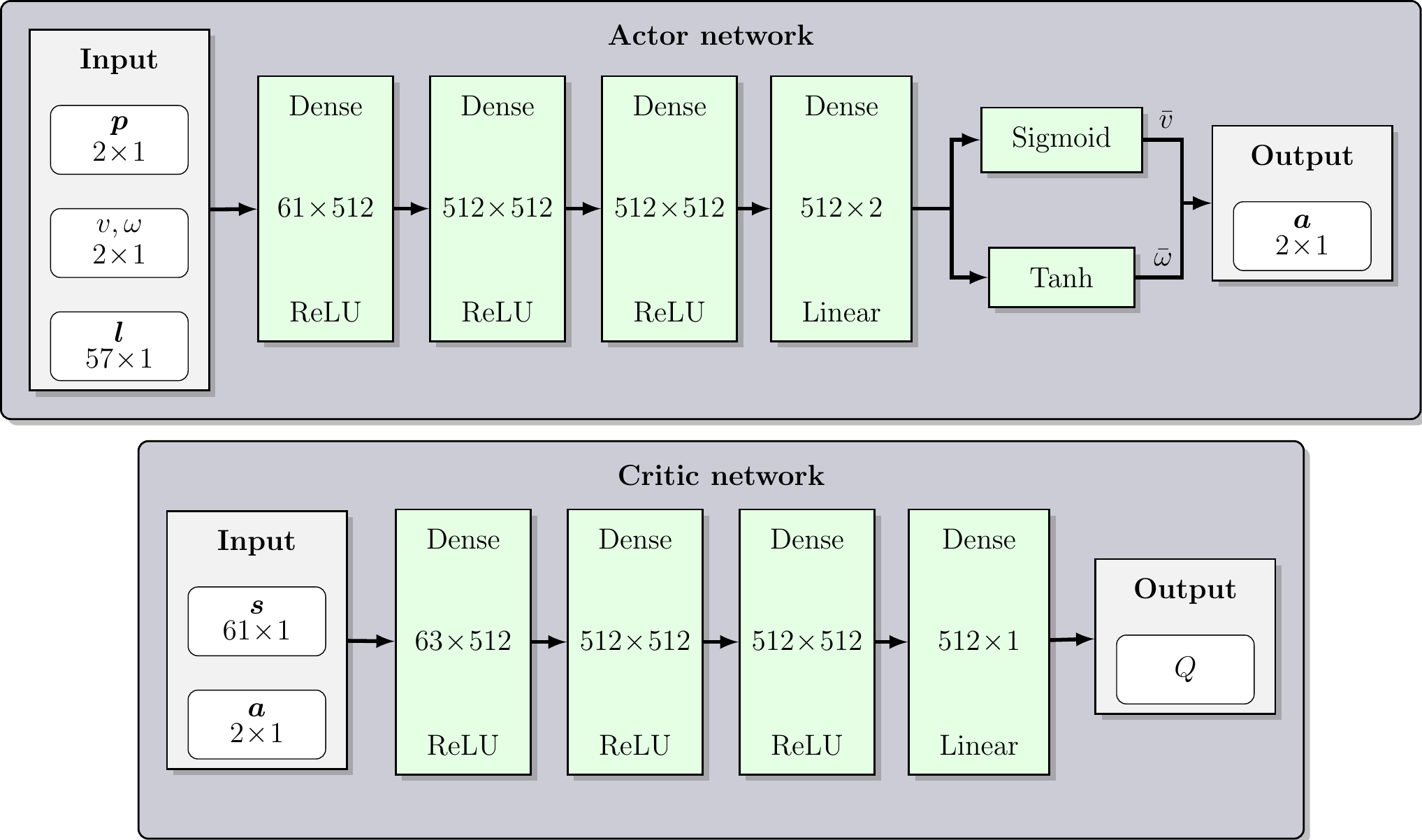}
    \caption{Actor network: input layer formed by the observation state space, followed by three dense ReLU layers of $512$ nodes, and the output action generated by merging values from a linear layer with a sigmoid and hyperbolic tangent activation functions.
    Critic network: input layer formed by the observation state space merged with the action space, followed by three dense ReLU layers of $512$ nodes and the output Q-value generated by a dense linear layer.}
    \label{fig:actor_proposed}
\end{figure}

The \ac{SAC} uses a double-$Q$ trick that requires two critic networks and one actor. Fig.~\ref{fig:actor_proposed} shows the proposed actor deep network structure composed of the observation state input layer, three hidden dense ReLU\footnote{Activation Function} layers with $512$ nodes each, and the action output generated by merging values from a linear layer with a sigmoid activation function for linear velocity and a hyperbolic tangent function for the angular velocity.

For the critic networks, the action generated by the actor is used as input merged to the observation state. As shown in Fig.~\ref{fig:actor_proposed}, the critic network structure also has three dense ReLU layers with $512$ nodes each, and the $Q$-value is generated by a linear activation function.
%
% \begin{figure}[t]
%     \centering
%      \includegraphics[trim={0in 0in 0in 0in},clip,width=1\textwidth]{images/actor.png}
%     \caption{Actor network: input layer formed by the observation state space, followed by three dense Relu layers of $512$ nodes, and the output action generated by merging values from a dense layer with a sigmoid and hyperbolic tangent activation functions.}
%     \label{fig:actor_proposed}
% \end{figure}
% %
% \begin{figure}[t]
%     \centering
%      \includegraphics[trim={0in 0in 0in 0in},clip,width=1\textwidth]{images/critic.png}
%     \caption{Critic network: input layer formed by the observation state space merged with the action space, followed by three dense Relu layers of $512$ nodes and the output Q-value generated by a dense linear layer.}
%     \label{fig:critic_proposed}
% \end{figure}

\section{Results}
\label{sec:results}

In this section, we evaluate our proposed approach by comparing it with the current literature. Two main aspects have been analyzed at this point: $i$) the impact of employing \ac{TD3} or \ac{SAC} algorithms in the context of safe local navigation for mobile robots, and $ii$) the effect of using our novel reward function instead others. To do so, we have first trained our policies in different simulated cluttered (sparse and complex) scenarios. Then, concerning the success rate (the number of times the robot reaches the goal region without collision in a limited time), we compare our method with other approaches in environments distinct from those used in the training stage. Next, to advance the generalization analysis of the solution, we added comparative trials in a second simulator (sim-to-sim analysis). Finally, we demonstrate the effectiveness of our proposal in a real-world cluttered scenario (sim-to-real analysis).

%%%%%%%%%%%%%%%%%%%%%%%%%%%%%%%%%%%
%%%%%%%%%%%%%%%%%%%%%%%%%%%%%%%%%%%
\subsection{Training setup}

\begin{figure}[t]
    \centering
    \subfigure[$12\times12$ m]{
         \includegraphics[clip,width=0.3\linewidth]{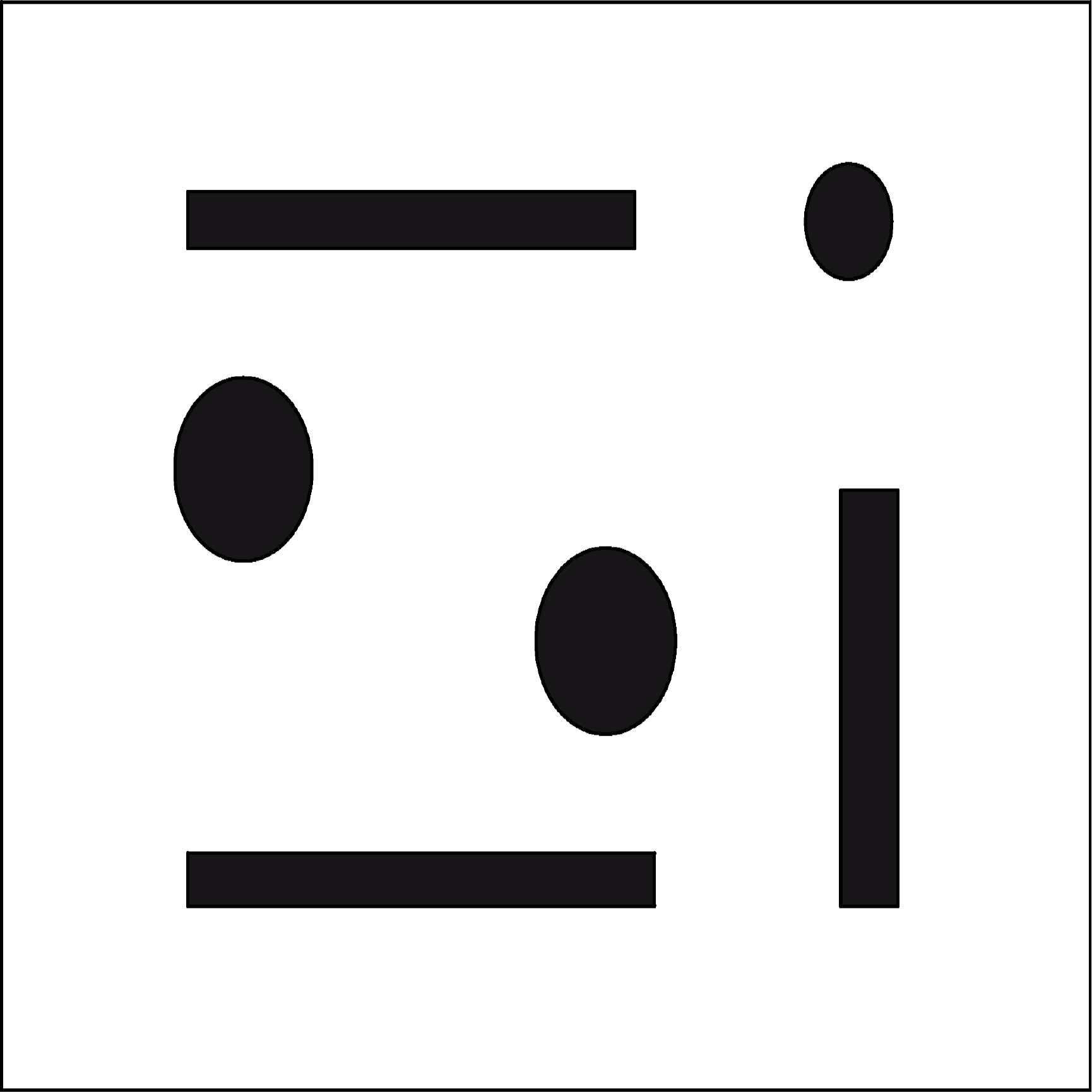}
         \label{fig:train_scenes_a}
     }
    \subfigure[$20\times20$ m]{
         \includegraphics[clip,width=0.3\linewidth]{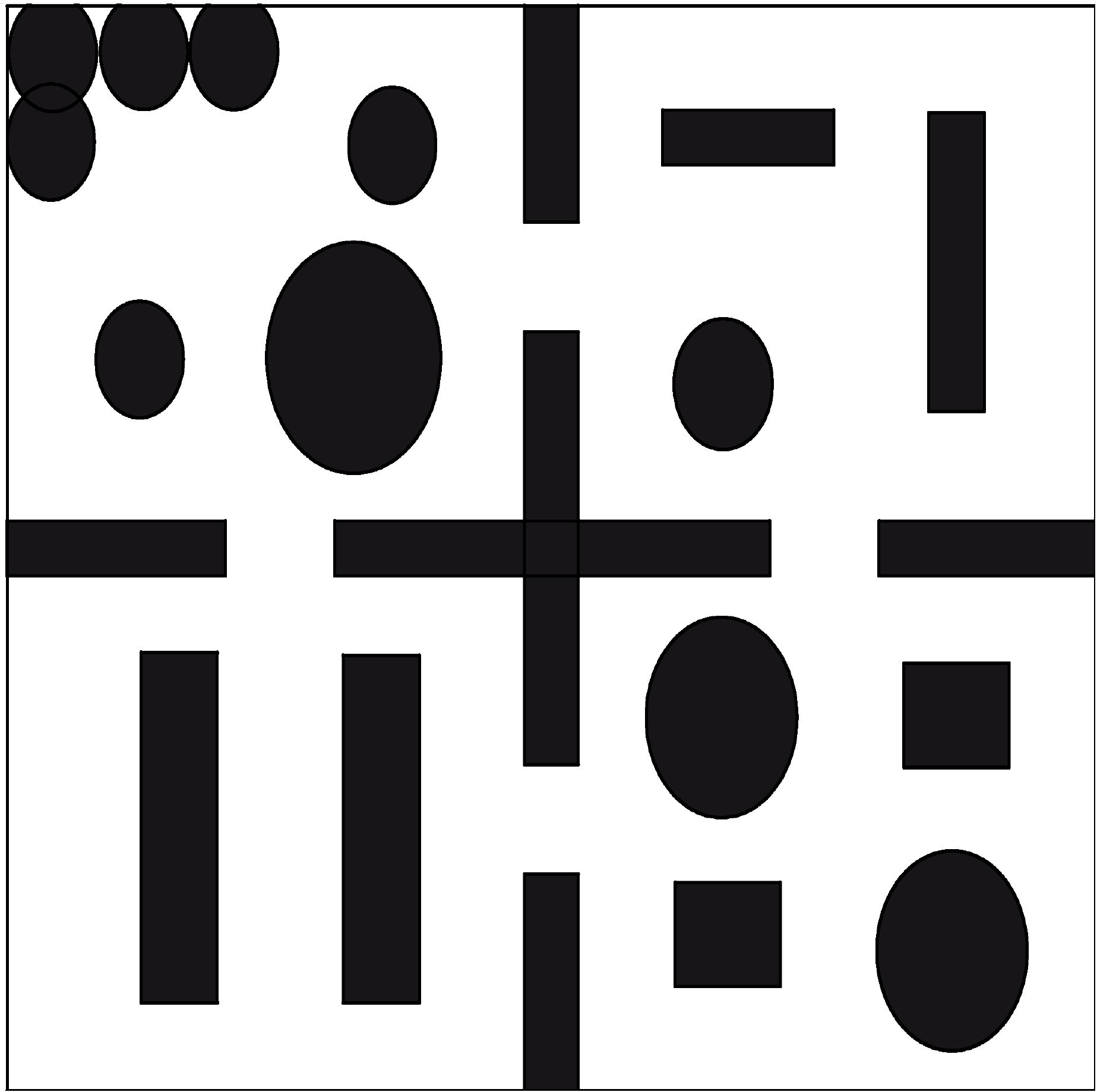}
         \label{fig:train_scenes_b}
    }
    \subfigure[$40\times40$ m]{
         \includegraphics[clip,width=0.3\linewidth]{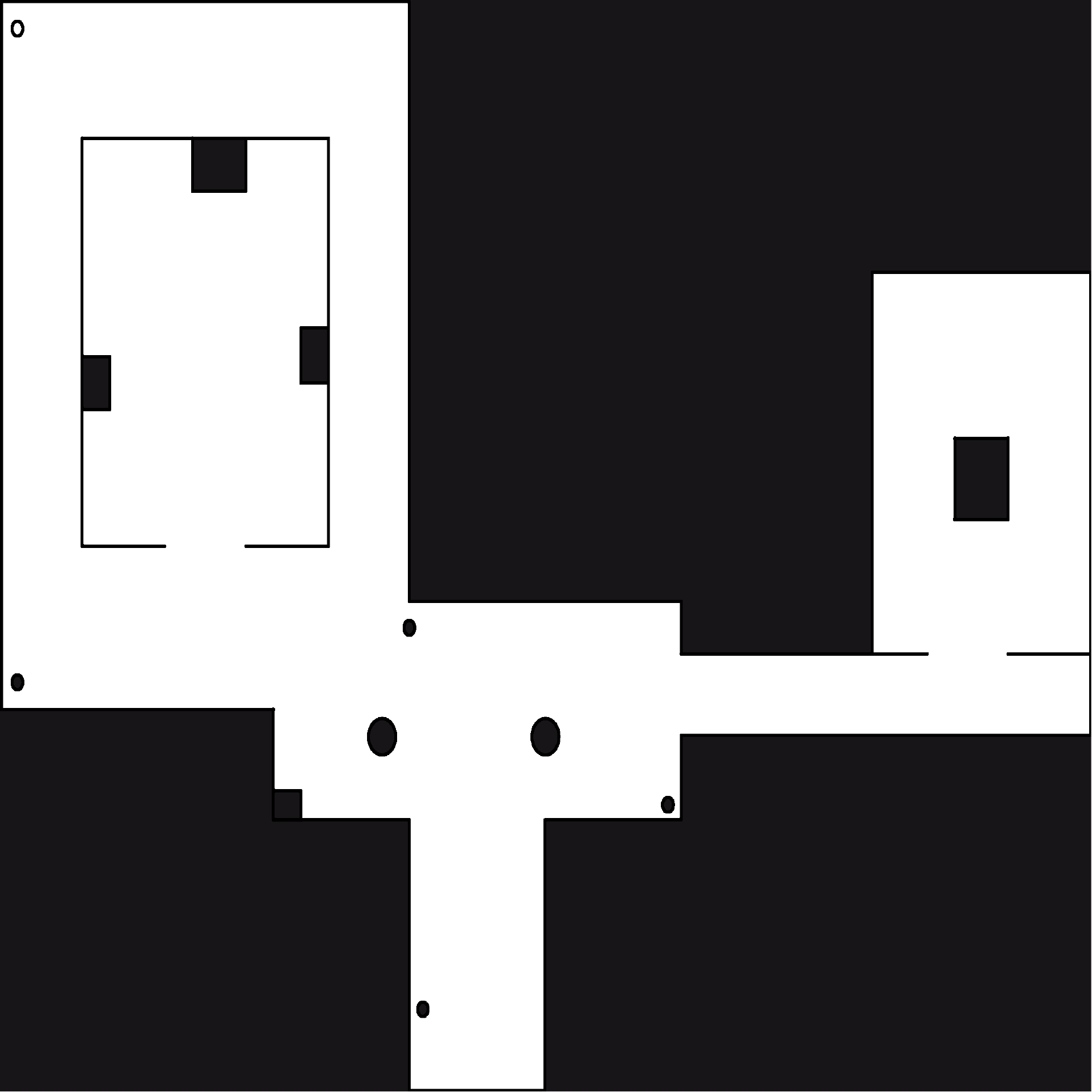}
         \label{fig:train_scenes_c}
    }
    \caption{Simulated scenarios used for training in learning step.}
    \label{fig:train_scenes}
\end{figure}

The training process has been performed on a laptop with Ubuntu 20.04, equipped with an NVIDIA GTX 3060 graphics card, 16 GB of RAM, and Intel Core i7-11800H CPU. The learning environment is represented by a differential wheeled robot navigating in limited workspaces, running in the standalone simulator presented in \cite{surmann2020deep}.
The three maps illustrated in Fig.~\ref{fig:train_scenes} have been employed at this step, whose dimensions are  $12\times12$ m, $20\times20$ m, and $40\times40$ m, respectively.

All policies used in the experiments were modeled according to the actor-critic deep networks illustrated in Fig.~\ref{fig:actor_proposed}, and they were trained by using the \ac{SAC} and the \ac{TD3} algorithms, both with 20000 episodes.
Each training episode ends when the robot reaches the goal region, collides with some obstacle, or violates the Timeout of 500 steps. The map changes randomly every 500 episodes among those shown in Fig.~\ref{fig:train_scenes}. Besides that, the actions was limited to $v \in [0,0.5]$ m/s and $\omega \in [-1,1]$ rad/s. Also, $d_{min} = 0.5$. The LiDAR has readings between $-135^\circ$ to $135^\circ$ with $684$ distance measurements. Seeking to reduce the state space for the training, we have divided these $684$ values into $57$ groups of $12$ measures and selected the 57 minimal distances of each group \cite{choi2020}.

Table \ref{tab:learning parameters} describes the learning parameters used for the training stage of the neural networks. In addition, we have set the constant reward values as: $r_a = 100$, $r_c = -200$ and $r_t = -200$. In the specific case of \ac{TD3}, the delayed rewards were updated over the last 10 steps. Additionally, from Eq.~\ref{eq:r_l}, $l_1 = 336$ and $l_2 = 348$, representing the LiDAR beam interval for the robot's front side read.
\begin{table}[h]
    \centering
    \caption{Learning parameters for \protect\ac{SAC} and \protect\ac{TD3}.}
     \label{tab:learning parameters}
      \resizebox{\textwidth}{!}{%
    \begin{tabular}{|c|c|c|c|c|}
    \hline
    \cellcolor{gray!20} \bf Learning Rate & \cellcolor{gray!20} \bf Batch Size & \cellcolor{gray!20} \bf Discount Factor & \cellcolor{gray!20} \bf Update Rate & \cellcolor{gray!20} \bf Policy Noise \\ \hline
    0.0003 & 256  & 0.99 & 0.005 & 0.2 \\
    \hline   
    \end{tabular}
    }
\end{table}

%%%%%%%%%%%%%%%%%%%%%%%%%%%%%%%%%%%
%%%%%%%%%%%%%%%%%%%%%%%%%%%%%%%%%%%
\subsection{Comparative Analysis}

\begin{figure}[t]
    \centering
    \subfigure[Map 1: $40\times40$ m]{
        \includegraphics[clip,width=0.3\linewidth]{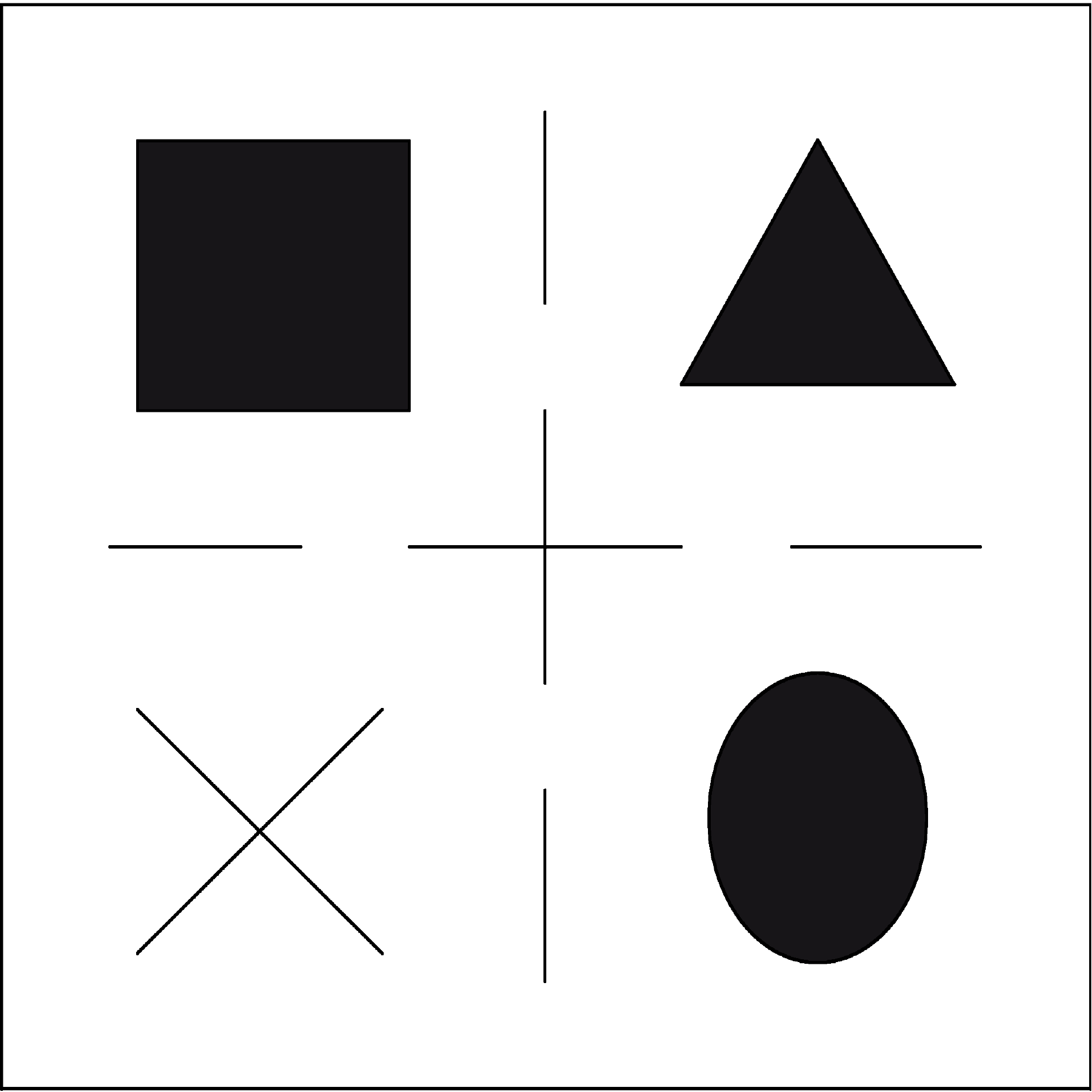}
        \label{fig:test_scenes_a}
    }
    %  \hfill
    \subfigure[Map 2: $40\times40$ m]{
        \includegraphics[clip,width=0.3\linewidth]{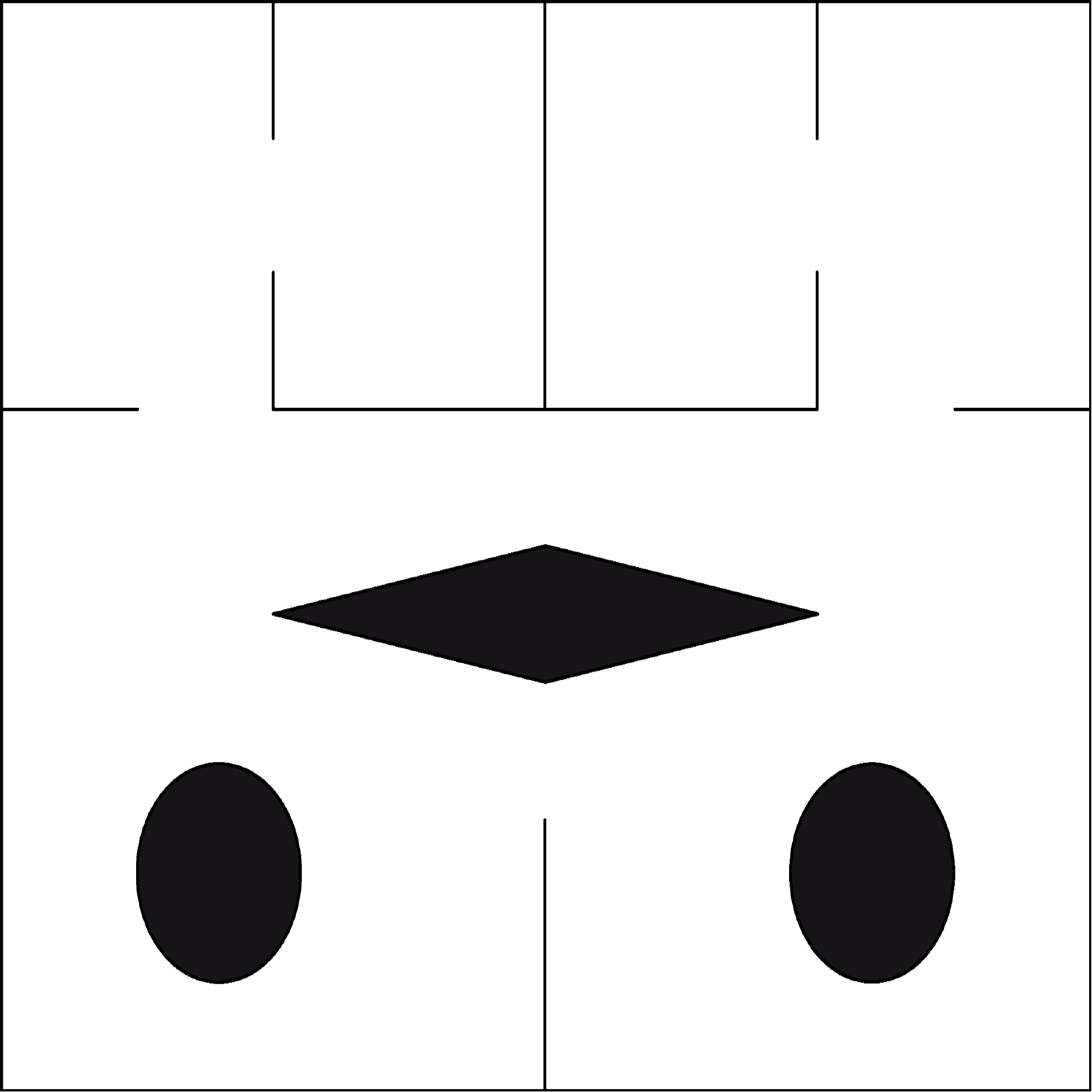}
        \label{fig:test_scenes_b}
    }
    %  \hfill
    \subfigure[Map 3: $60\times60$ m]{
        \includegraphics[clip,width=0.3\linewidth]{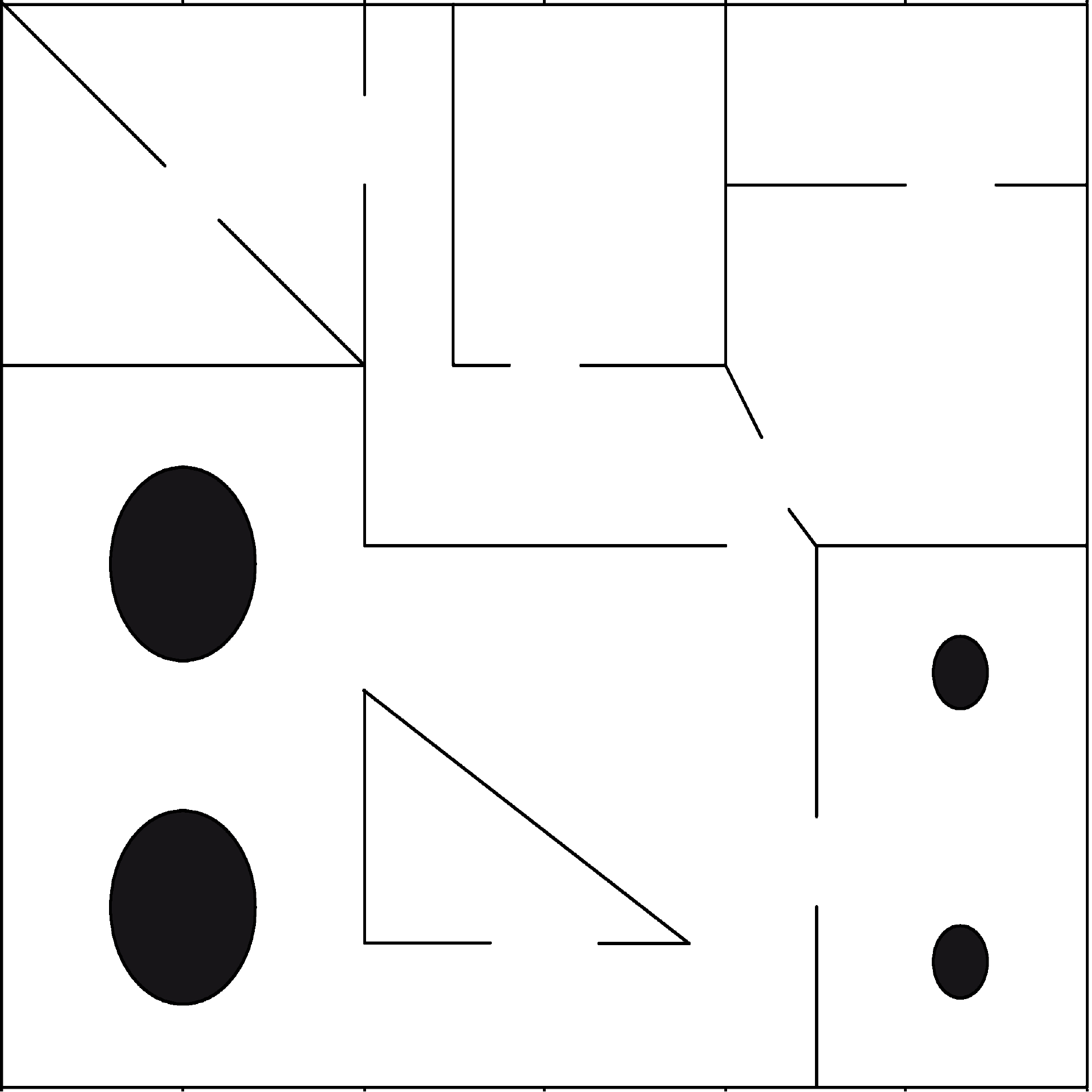}
        \label{fig:test_scenes_c}
    }
    \caption{Simulated scenarios used for testing and evaluating the algorithms learning performance.}
    \label{fig:test_scenes}
\end{figure}

Our main goal in this paper is to evaluate how generalizable our technique is when compared to other existing ones. Therefore, after obtaining the trained policies, we execute some tests to analyze how they perform in scenarios distinct from those previously trained. Fig.~\ref{fig:test_scenes} illustrates the tested maps, relatively larger than those of Fig.~\ref{fig:train_scenes}, with dimensions $40\times40$ m, $40\times40$ m, and $60\times60$ m, respectively.

In our comparative analysis, we have used three different reward functions from the state-of-the-art literature. In \cite{cimurs2021goal}, the reward is based only on the robot's linear and rotational speeds, its distance to the goal, and collision information.
In \cite{hu2020voronoi}, the authors also used the robot's speeds and distance to the goal, but they incorporates exteroceptive (\ac{LiDAR}) data to compute a safety clearance reward term for collision avoidance purposes.
Finally, in \cite{grando2021deep}, the reward is the simplest one, using only collision information and the arrival at the goal.

Our trials were separated into two groups. First, for all rewards functions (including ours), we have applied the \ac{TD3} algorithm proposed in \cite{fujimoto2018addressing} to train them. Here it is important to highlight that \cite{cimurs2021goal} have used the \ac{TD3} to learn the navigation policy, while \cite{hu2020voronoi} and \cite{grando2021deep} used the \ac{DDPG} \cite{lillicrap2015continuous}, a predecessor of the \ac{TD3} and \ac{SAC}. Second, for the same reward functions, we have replaced  \ac{TD3} by \ac{SAC} algorithm.

The learning parameters used for training all structures were the same in Table~\ref{tab:learning parameters}. However, as we have considered different reward functions for comparing, the reward constants received values according to the author's proposal of each paper in analysis, as shown in Table~\ref{tab:literature_rewards}.
\begin{table}[h]
    \centering
    \caption{Policy reward constant values used by literature papers considered for comparing.}
     \label{tab:literature_rewards}
      \resizebox{\textwidth}{!}{%
    \begin{tabular}{|c|c|c|c|}
    \hline
\cellcolor{gray!20} \bf Reward/Reference & \cellcolor{gray!20} \bf Strategy-\cite{cimurs2021goal} & \cellcolor{gray!20} \bf Strategy-\cite{hu2020voronoi} & \cellcolor{gray!20} \bf Strategy-\cite{grando2021deep} 
    \\ \hline
    Collision & $r_c = -100$  & - & $r_c = -10$ \\
    Arrival & $r_a = 80$  & $r_a = 40$ & $r_a = 100$ \\
    Others & - & $r_{cp}=r_{cpo}=r_{av}=r_{lv}= -1$ & - \\
    \hline   
    \end{tabular}
    }
\end{table}

\begin{figure}[t]
    \centering
    \subfigure[Reward function in \cite{cimurs2021goal}.]{
        \includegraphics[width=0.47\linewidth]{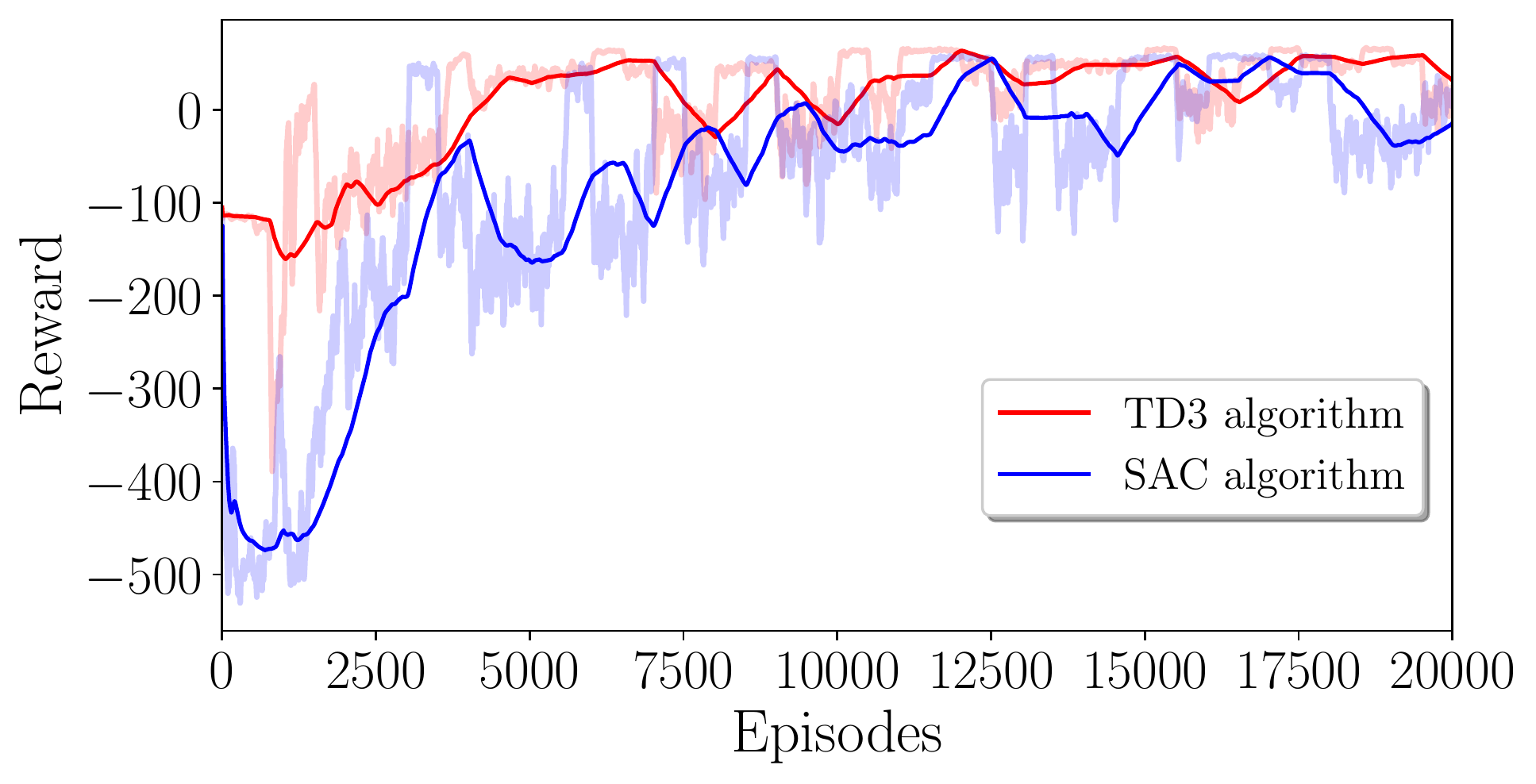}
        \label{fig:rewards_8}
    }
    \subfigure[Reward function in \cite{hu2020voronoi}.]{
        \includegraphics[width=0.47\linewidth]{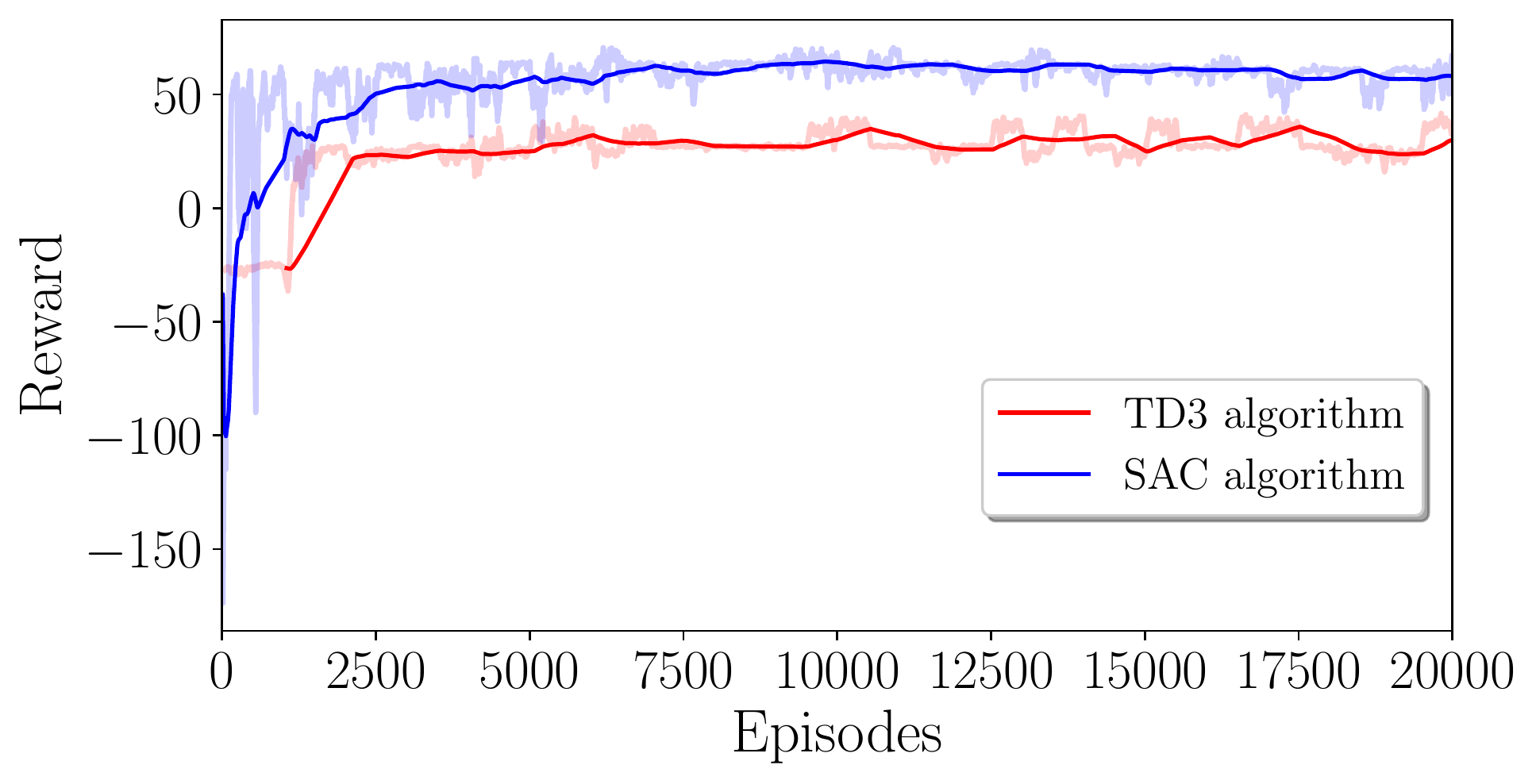}
        \label{fig:rewards_9}
    }
    \subfigure[Reward function in \cite{grando2021deep}.]{
        \includegraphics[width=0.47\linewidth]{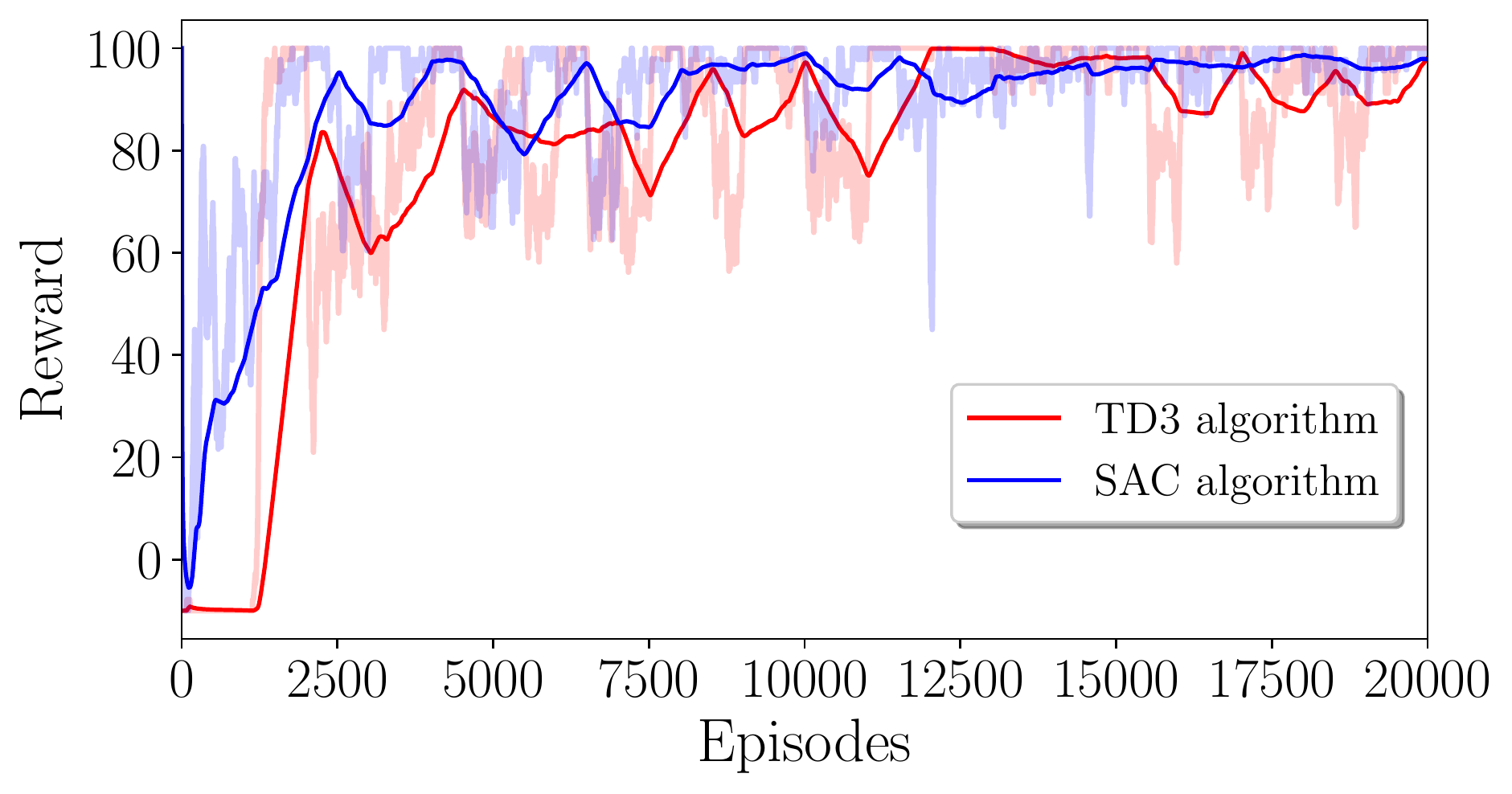}
        \label{fig:rewards_10}
    }
    \subfigure[Our reward function.]{
        \includegraphics[width=0.47\linewidth]{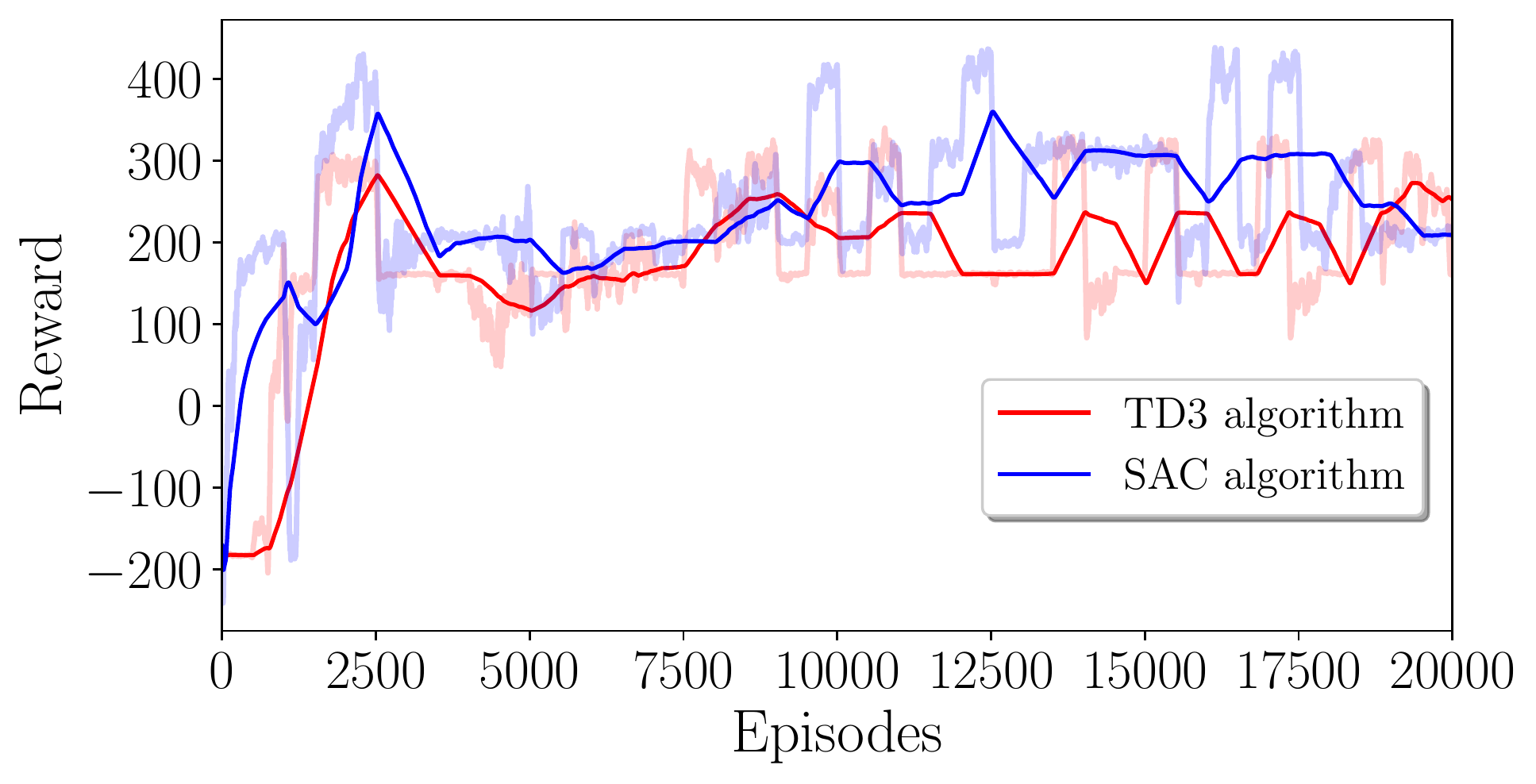}
        \label{fig:rewards_ours}
    }
    \caption{Moving average of all compared reward functions for the \protect\ac{TD3} (in red) and the \protect\ac{SAC} (in blue) algorithms.}
    \label{fig:rewards}
\end{figure}

Fig.~\ref{fig:rewards} presents the moving average evolution for all aforementioned reward functions, each one trained with both, \ac{TD3} and \ac{SAC} algorithms. In most cases, the rewards evolve side by side (with some variations) throughout the training. The periodic oscillations that appear in the graphs refer to each change in the training map, where the value decreases for a while and increases according to the train evolution in the new map.

Table \ref{tab:simulation_experiments_TD3} compiles the results obtained for the first set of simulations using the \ac{TD3}. The percentage value for each entry \emph{scenario/method} was calculated by counting the number of trials in which the robot was capable of reaching the goal position before the timeout of 1500 steps or colliding with an obstacle after 500 attempts. For each attempt, we have set the starting and goal locations following 15 pairs of positions that were repeated during all experiments. 
The first observation is that our reward function outperforms all others in terms of completed missions. The results were significantly superior, although \cite{hu2020voronoi} also stood out from the other two. It can possibly be explained by the fact that, with the use of a safety clearance reward, the agent learns better how to avoid collisions and local minima. Additionally, the use of perception information about the increasing of the map (exteroceptive information) helped to deal with local minima problems.

\begin{table}[htpb]
    \centering
    \caption{Performance comparison among our proposed reward and others from the literature using the \protect\ac{TD3} algorithm on untrained scenarios.}
     \label{tab:simulation_experiments_TD3}
    %
    % \resizebox{0.98\textwidth}{!}{%
    \begin{tabular}{|l|c|c|c|c|}
    \hline
    \cellcolor{gray!20} \bf Scenario/Method & \cellcolor{gray!20} \bf TD3-\cite{cimurs2021goal} & \cellcolor{gray!20} \bf TD3-\cite{hu2020voronoi} & \cellcolor{gray!20} \bf TD3-\cite{grando2021deep} & \cellcolor{gray!20} \bf TD3-Ours \\ \hline
    Map 1 -- Fig.\ref{fig:test_scenes_a} & 63.4\%  & 81.6\% & 56.2\% & \bf 86.4\% \\
    \hline   
    Map 2 -- Fig.\ref{fig:test_scenes_b} & 51.4\% & 54.2\% & 36.6\% & \bf 57.2\% \\
    \hline   
    Map 3 -- Fig.\ref{fig:test_scenes_c} & 18.2\% & 38.4\% & 33.8\% & \bf 42.0\% \\
    \hline           
    \end{tabular}
    % }
\end{table}

The results also allow us to conclude that Map 1 (Fig.\ref{fig:test_scenes_a}) is the least complex environment, while Map 3 (Fig.\ref{fig:test_scenes_c}) is the most complex regarding obstacles and local minima situations, once all reward functions follow this pattern. 
Still evaluating the reward functions, an outlier appears in scenario 3 when solely comparing the performance obtained by the rewards presented in \cite{cimurs2021goal} and \cite{grando2021deep}. Although \cite{cimurs2021goal} is better in maps 1 and 2, the performance deteriorates sharply in map 3 even with a simpler reward function as the one proposed in \cite{grando2021deep}.

Table \ref{tab:simulation_experiments_SAC} contains the results obtained from tests performed using the \ac{SAC}, and following the same test parameters used for \ac{TD3}. Results show an increase in the number of successfully completed trials when using \ac{SAC} for training compared to \ac{TD3}. This performance shows that, for the robot navigation policies, increasing the exploration of actions during training helps the agent to select better actions in untrained scenarios, that is, increasing the generalization of the trained network. Therefore, adopting exploration techniques as including the entropy maximization on training, which is done in the \ac{SAC} algorithm, helps to increase the generalization.

Although most of the results present this performance increase comparing the model trained using SAC instead of TD3, the proposed rewards functions presented in \cite{hu2020voronoi} and \cite{grando2021deep} decreased only in scenario 3. However, these results can be observed as outliers since, in all other experiments, the SAC results were better than TD3, and also maintaining the performance results observed about the different rewards function.

%%%
%
\begin{table}[h]
    \centering
    \caption{Performance comparison among our proposed reward and others from the literature using the \protect\ac{SAC} algorithm on untrained scenarios.}
    \label{tab:simulation_experiments_SAC}
    %
    % \resizebox{0.98\textwidth}{!}{%
    \begin{tabular}{|l|c|c|c|c|}
    \hline
    \cellcolor{gray!20} \bf Scenario/Method & \cellcolor{gray!20} \bf SAC-\cite{cimurs2021goal} & \cellcolor{gray!20} \bf SAC-\cite{hu2020voronoi} & \cellcolor{gray!20} \bf SAC-\cite{grando2021deep} & \cellcolor{gray!20} \bf SAC-Ours \\ \hline
    Map 1 -- Fig.\ref{fig:test_scenes_a} & 80.2\% & 88.8\% & 64.8\% & \bf 95.2\% \\
    \hline   
    Map 2 -- Fig.\ref{fig:test_scenes_b} & 61.0\% & 61.6\% & 45.2\% & \bf 83.0\% \\
    \hline   
    Map 3 -- Fig.\ref{fig:test_scenes_c} & 35.6\% & 34.2\% & 31.6\% & \bf 59.4\% \\
    \hline           
    \end{tabular}
    % }
\end{table}

Regarding the proposed reward function some benefits can be commented. Even when using the TD3 algorithm, our approach outperforms SAC-\cite{cimurs2021goal} and SAC-\cite{grando2021deep}, for map 1; SAC-\cite{grando2021deep} for map 2; and all methods for map 3. It is slightly worse than SAC-\cite{hu2020voronoi}, for maps 1 and 2 and (under 4.5\% difference), and than SAC-\cite{cimurs2021goal}, for map 2 (under 4\% difference).

When the best combinations of reward and training algorithm are compared, our approach outperforms all competitors. Comparing to the second best method, it is 7.2\%, 34.7\% and 54.7\% for maps 1 to 3, respectively.

%%%%%%%%%%%%%%%%%%%%%%%%%%%%%%%%%%%
%%%%%%%%%%%%%%%%%%%%%%%%%%%%%%%%%%%
\subsection{Sim-to-sim analysis}

Sim-to-sim analysis is often an intermediate step to sim-to-real transfer tests since, in Robotics, collecting and validating data in the real world is more costly \cite{Stephen2019Sim}.
Although we have demonstrated in the previous section that our approach is more generalizable than others in the current literature in the context of robot navigation in cluttered environments, it has been done using the standalone simulator presented in \cite{surmann2020deep}, which is more simple in terms of dynamics, friction effects, and disturbances in general. Our choice was based on the fact that a simpler simulator tends to be more efficient in the training stage, allowing greater exploration of the environment in a shorter time.

Therefore, in this section, we present a comparative analysis concerning a more realistic simulator, the \emph{CoppeliaSim}\footnote{https://www.coppeliarobotics.com/}, integrated with \ac{ROS} Noetic in a laptop with Ubuntu 20.04, NVIDIA GTX 3060 graphics card, 16 GB of RAM, and Intel Core i7-11800H CPU. The simulations rely on a Pioneer P3dx equipped with a Velodyne VPL-16 but reading only a 2D scan from the 3D point cloud. The 2D scans also have readings between $-135^\circ$ to $135^\circ$ with $684$ distance measurements and decimated in $57$ values of minimum distances detected, as described in Section~\ref{subsec:obs_statesDescription}.

\begin{figure}[t]
    \centering
    \subfigure[$40\times30$ m]{
         \includegraphics[width=0.46\textwidth]{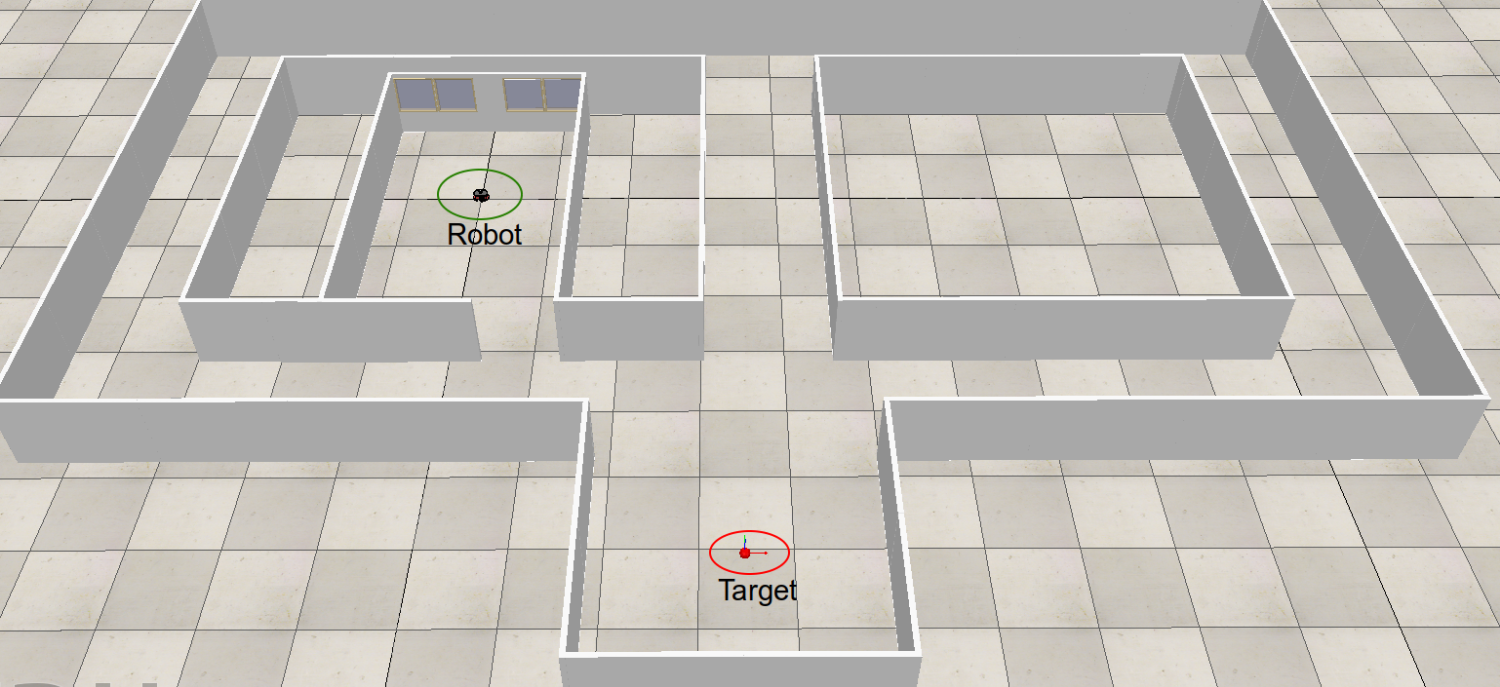}
         \label{fig:coppelia_scenes_a}
     }
    \subfigure[$40\times30$ m]{
         \includegraphics[width=0.46\textwidth]{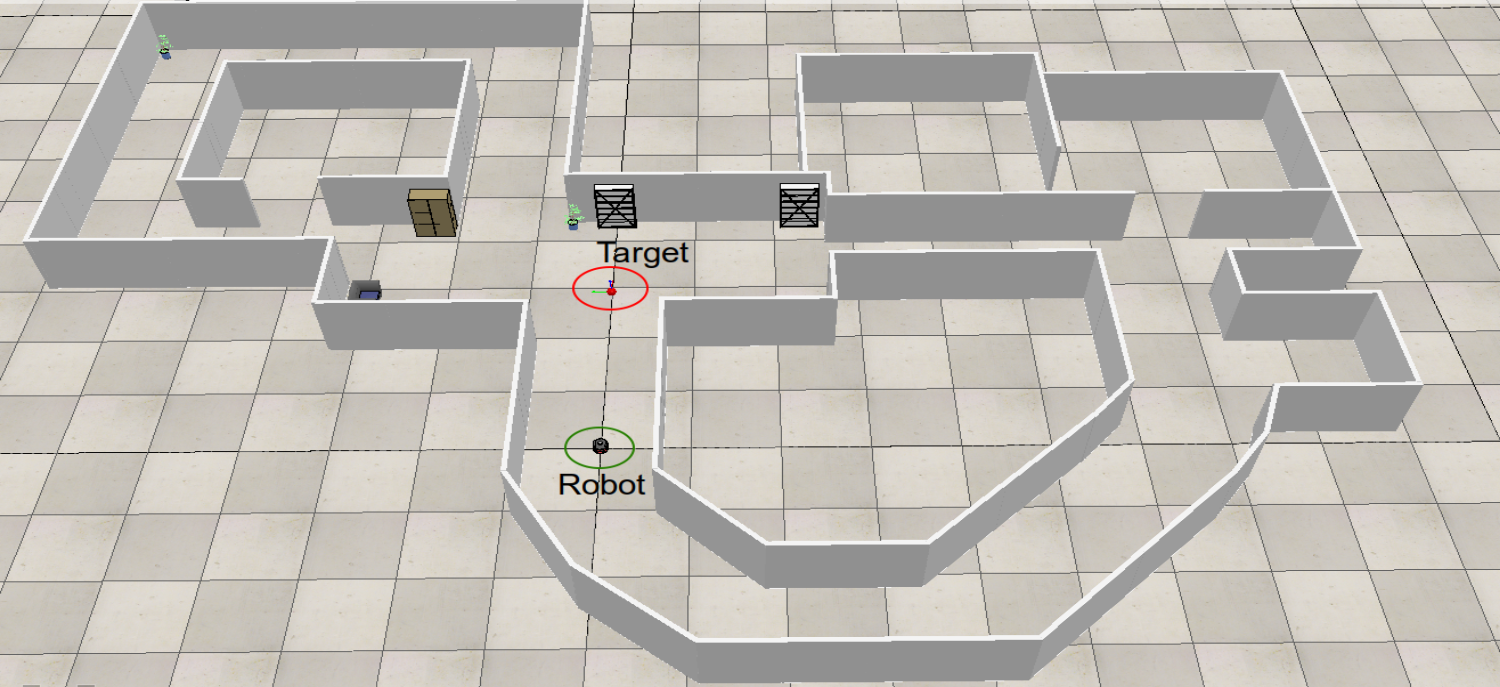}
         \label{fig:coppelia_scenes_b}
    }
    \caption{Simulated scenarios used for the test in CoppeliaSim.}
    \label{fig:coppelia_scenes}
\end{figure}

Here, the set of trials was performed in the two environments illustrated in Fig.~\ref{fig:coppelia_scenes}, both with dimensions $40\times30$ m. In the last section, the reward function in \cite{hu2020voronoi} shows to be the second most efficient, then we incorporate it in this new test in two versions, one with the \ac{TD3} algorithm and another with the \ac{SAC}. Also, we have limited our proposed method to the \ac{SAC} method, which presented a better performance in the previous section.

Table \ref{tab:simulation_experiments_Coppelia} presents the success rates for the simulations in the CoppeliaSim, using the same policies trained in the previous section. This time, due to the higher cost of running the simulator, we executed 200 trials for each table entry.
As it can be seen, our method also outperforms \cite{hu2020voronoi}, both with the \ac{TD3} and the \ac{SAC}, in the two specific scenarios.

%%% Tabela Coppelia
%
\begin{table}[htpb]
    \centering
    \label{tab:simulation_experiments_Coppelia}
    \caption{Performance comparison between the proposed learning structure and the literature best results obtained using SAC and TD3 in CoppeliaSim simulator scenarios.}
    %
    % \resizebox{0.98\textwidth}{!}{%
    \begin{tabular}{|l|c|c|c|}
        \hline
        \cellcolor{gray!20} \bf Scenario/Method & \cellcolor{gray!20} \bf TD3-\cite{hu2020voronoi} & \cellcolor{gray!20} \bf SAC-\cite{hu2020voronoi} & \cellcolor{gray!20} \bf SAC-Ours \\ \hline
        Coppelia map 1 -- Fig.\ref{fig:coppelia_scenes_a} & 75.0\% & 83.5\% & \bf 91.0\% \\
        \hline   
        Coppelia map 2 -- Fig.\ref{fig:coppelia_scenes_b}& 72.5\% & 76.5\%  & \bf 92.5\% \\
        \hline           
        \end{tabular}
    % }
\end{table}

Concerning the binomial probability for the first map (Fig.\ref{fig:coppelia_scenes_a}) the \ac{SAC}-\cite{hu2020voronoi} presents 90\% confidence interval of approximately [78\%, 87\%], while our method falls between [87\%, 94\%]. Then, our method has a 90\% chance of being better than the best one in the current literature. Results are even better for the second map (Fig.\ref{fig:coppelia_scenes_b}) where the 99\% confidence interval is [67\%, 83\%] for the \ac{SAC}-\cite{hu2020voronoi}, and [88\%, 94\%] for our method.
The main conclusion at this point is that our proposed approach was capable of dealing better with the differences between the previous simulator (in which all policies have been trained) and the CoppeliaSim, at least for the used maps.

% Map 1
% Deles
% 99\% confidence interval:	0.75141 ≤ p ≤ 0.89233
% 95\% confidence interval:	0.77063 ≤ p ≤ 0.87930
% 90\% confidence interval:	0.78020 ≤ p ≤ 0.87229
% Nosso
% 99\% confidence interval:	0.84519 ≤ p ≤ 0.95435
% 95\% confidence interval:	0.86149 ≤ p ≤ 0.94579
% 90\% confidence interval:	0.86946 ≤ p ≤ 0.94102

% Map 2
% Deles
% 99% confidence interval:	0.67400 ≤ p ≤ 0.83325
% 95% confidence interval:	0.69469 ≤ p ≤ 0.81743
% 90% confidence interval:	0.70510 ≤ p ≤ 0.80903
% Nosso
% 99% confidence interval:	0.85754 ≤ p ≤ 0.96144
% 95% confidence interval:	0.87334 ≤ p ≤ 0.95358
% 90% confidence interval:	0.88103 ≤ p ≤ 0.94918

%%%%%%%%%%%%%%%%%%%%%%%%%%%%%%%%%%%
%%%%%%%%%%%%%%%%%%%%%%%%%%%%%%%%%%%
\subsection{Sim-to-real experiments}

Finally, to evaluate our proposed method, we have applied it to navigate a robot in the real world. Here, the main idea is to qualitatively evaluate the \emph{sim-to-real} transfer capacity, which are concrete instances of the generalization problem \cite{kirk2021survey}. Therefore, we didn't perform a comparison with other methods addressed in previous experiments.

The experiment consists of navigating through an environment formed by corridors to a target point, avoiding obstacles. The robot selected for the task was a \emph{Pioneer 3at} equipped with a \emph{Hokuyo UTM-30LX-EW} \ac{LiDAR}, an intel Realsense T267 for localization, and a \emph{Jetson TX2} NVIDIA Pascal GPU architecture with 256 CUDA cores, Ubuntu 18.04 and \ac{ROS} Melodic.

\begin{figure}[htpb]
    \centering
     \begin{subfigure}
         \centering
         \includegraphics[trim={0.0in 0.0in 0.0in 0.0in},width=0.318\textwidth]{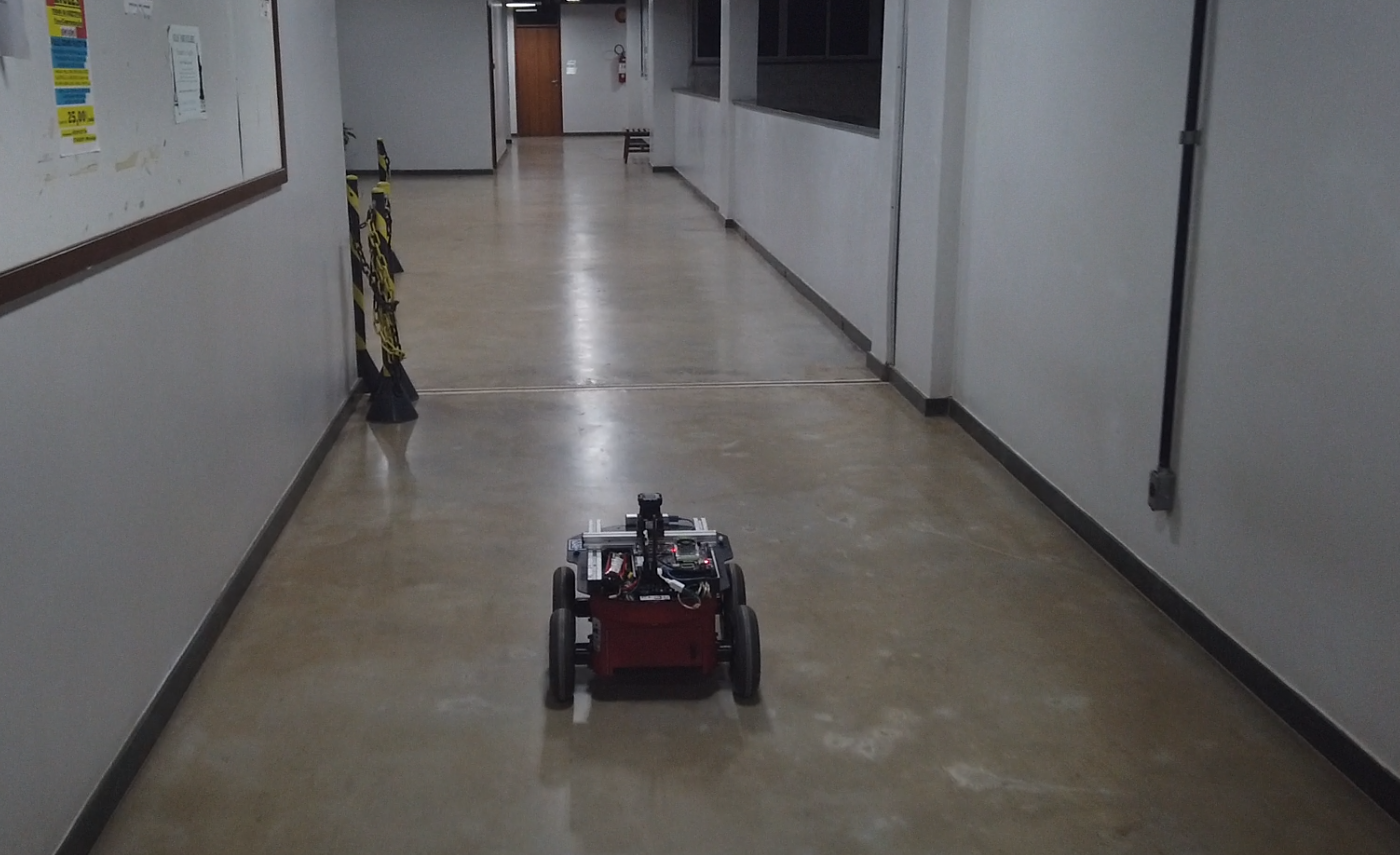}
         \label{fig:video_frame_a}
     \end{subfigure}
    %  \hfill
     \begin{subfigure}
         \centering
         \includegraphics[trim={0cm 0.0cm 0cm 0cm},clip,width=0.318\textwidth]{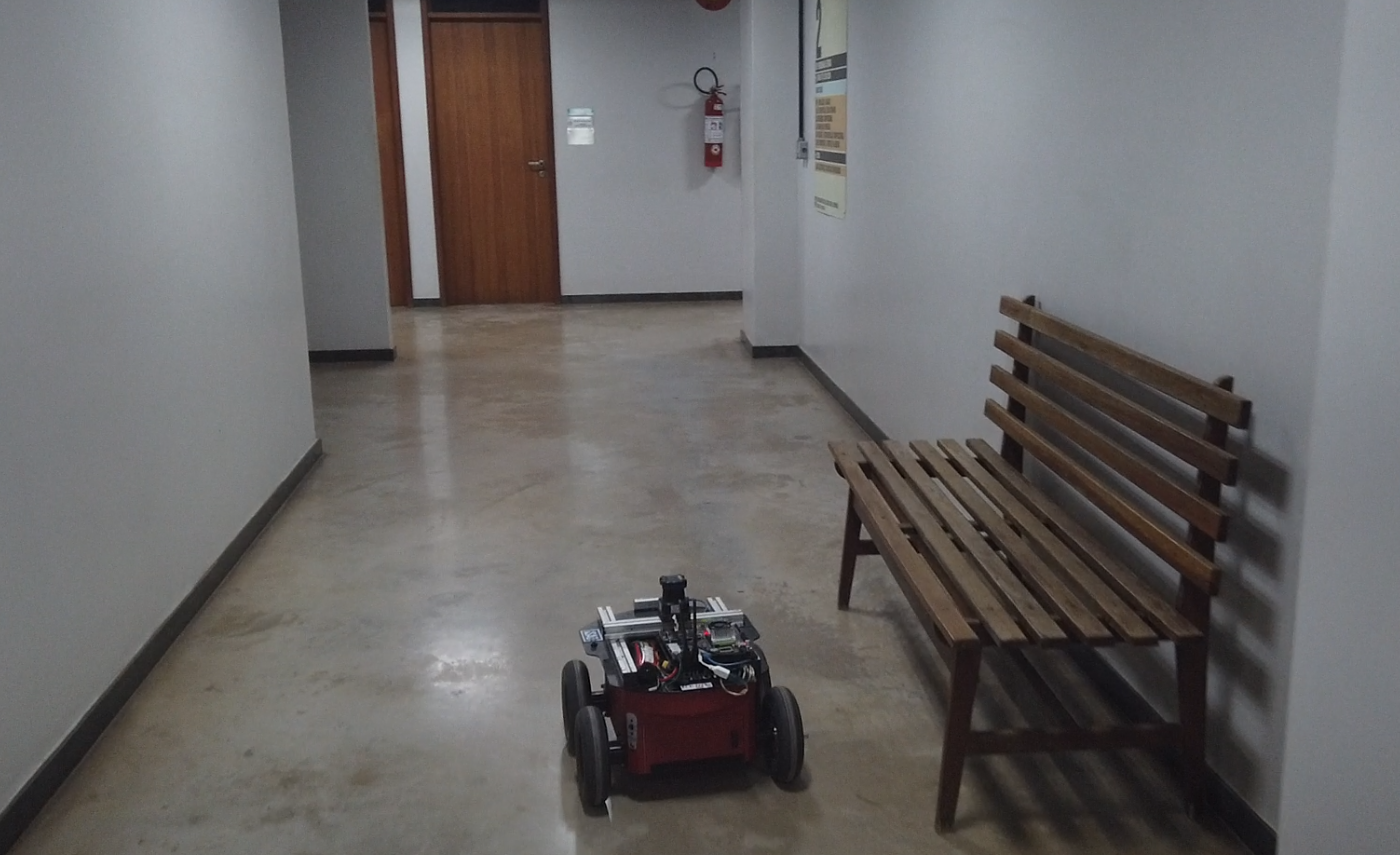}
         \label{fig:video_frame_b}
     \end{subfigure}
    %  \hfill
     \begin{subfigure}
         \centering
         \includegraphics[trim={0cm 0.0cm 0cm 0cm},clip,width=0.318\textwidth]{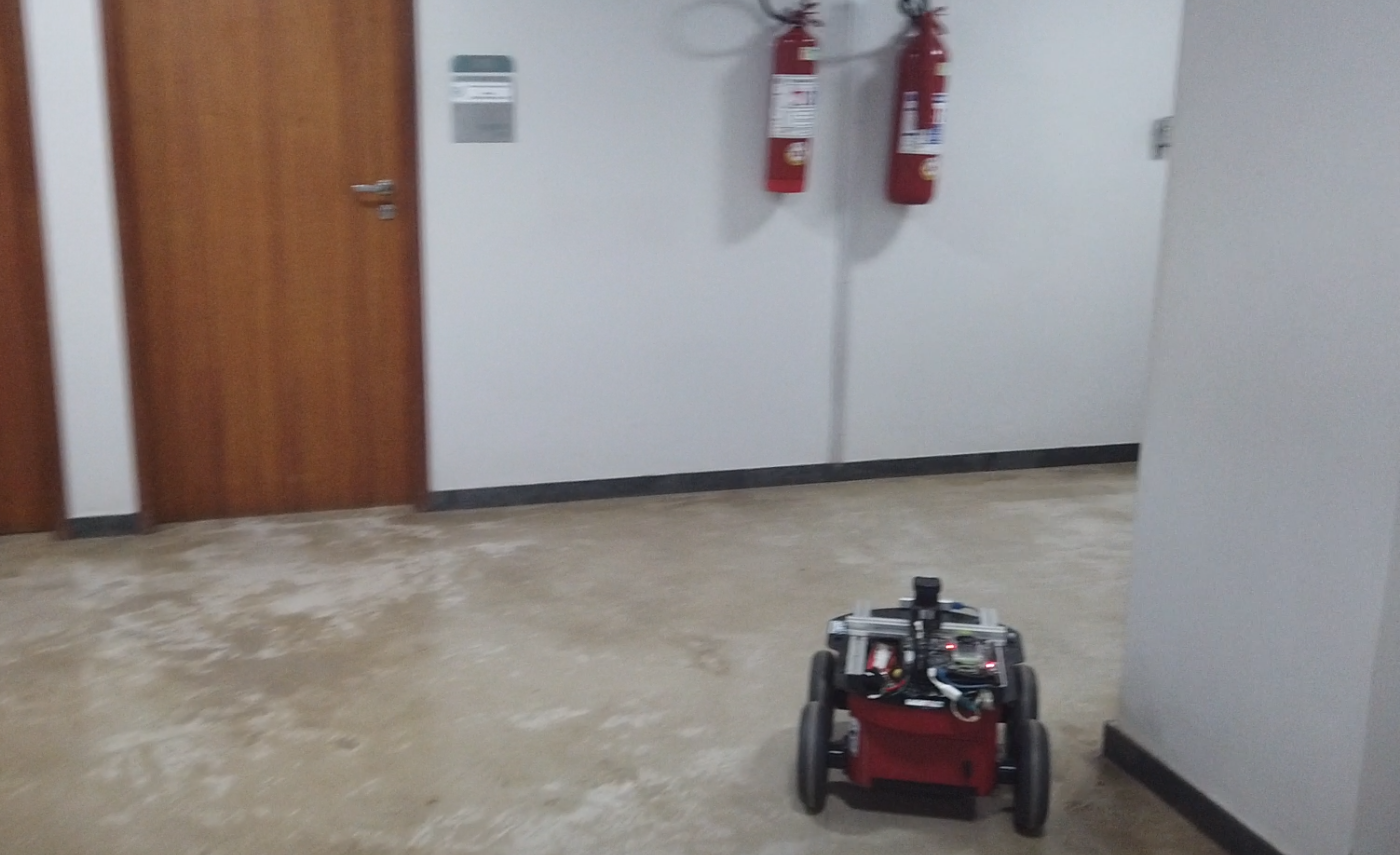}
         \label{fig:video_frame_c}
     \end{subfigure}
    %  \hfill
     \begin{subfigure}
         \centering
         \includegraphics[trim={0.0in 0.0in 0.0in 0.0in},width=0.318\textwidth]{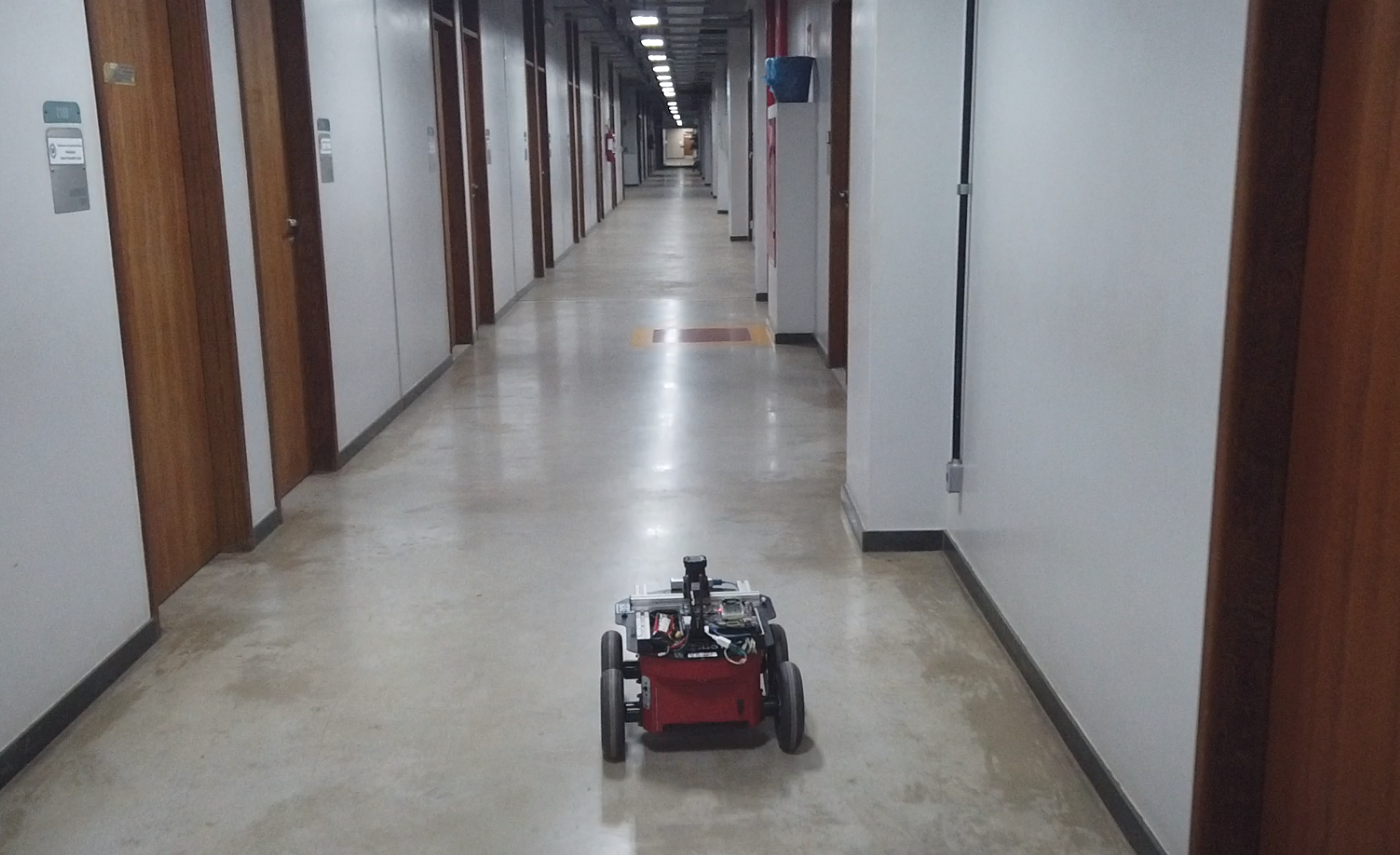}
         \label{fig:video_frame_d}
     \end{subfigure}
    %  \hfill
     \begin{subfigure}
         \centering
         \includegraphics[trim={0cm 0.0cm 0cm 0cm},clip,width=0.318\textwidth]{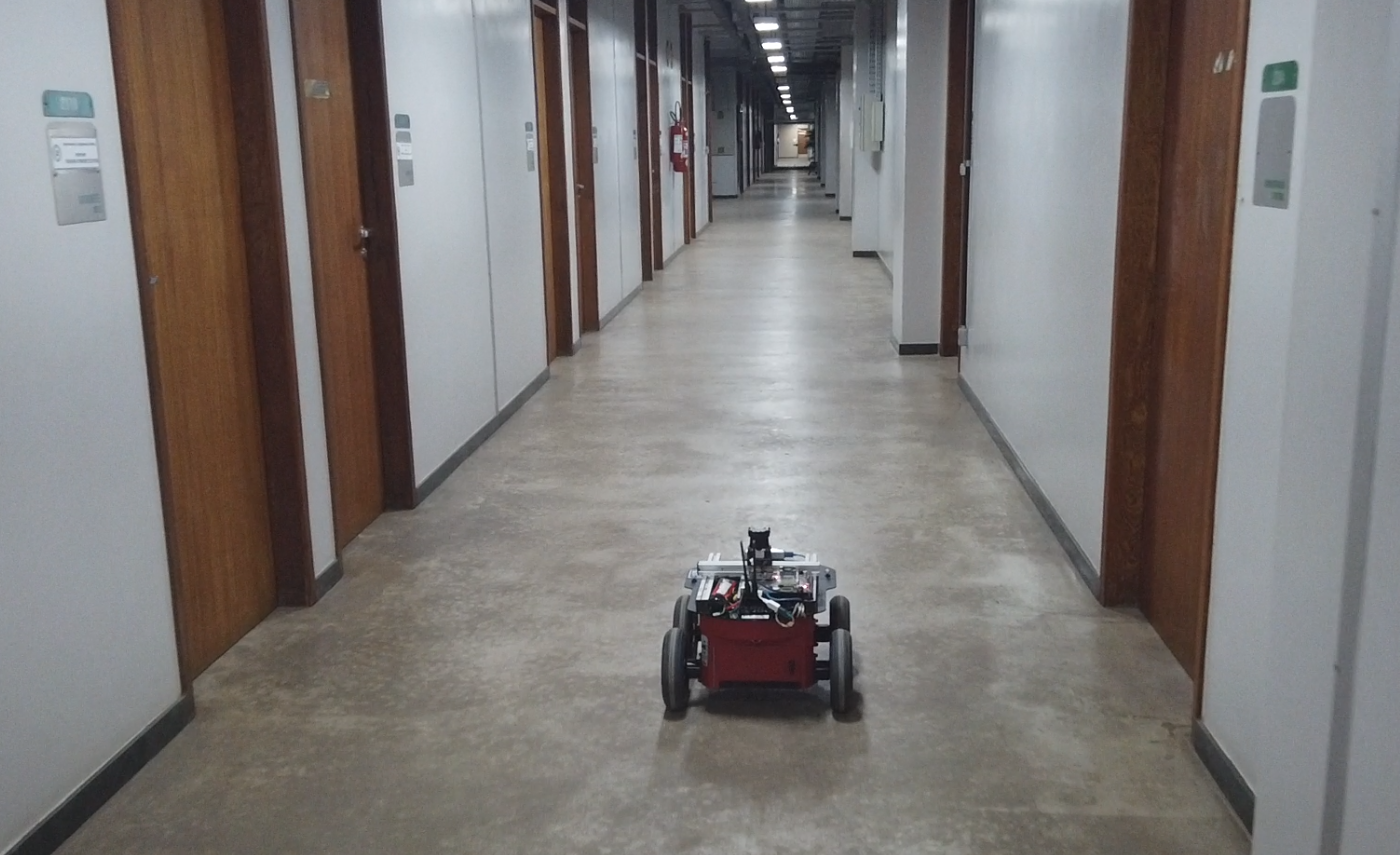}
         \label{fig:video_frame_e}
     \end{subfigure}
    %  \hfill
     \begin{subfigure}
         \centering
         \includegraphics[trim={0cm 0.0cm 0cm 0cm},clip,width=0.318\textwidth]{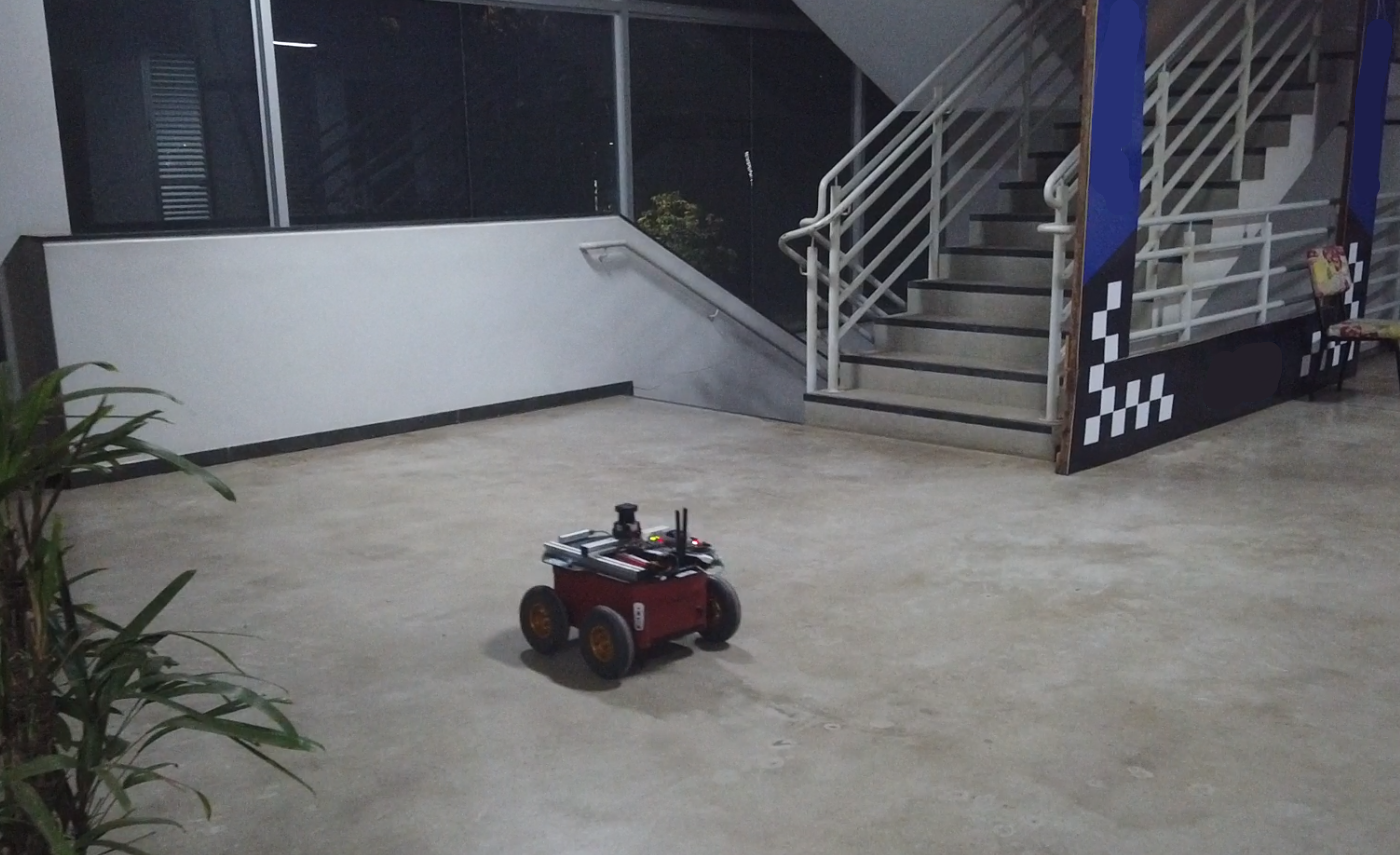}
         \label{fig:video_frame_f}
     \end{subfigure}
    \caption{Frame sequence of the real-world experiment.}
    \label{fig:video_frame}
\end{figure}

\begin{figure}[t]
    \centering
    \includegraphics[trim={0cm 0.0cm 0cm 0.7cm},clip,width=1\linewidth]{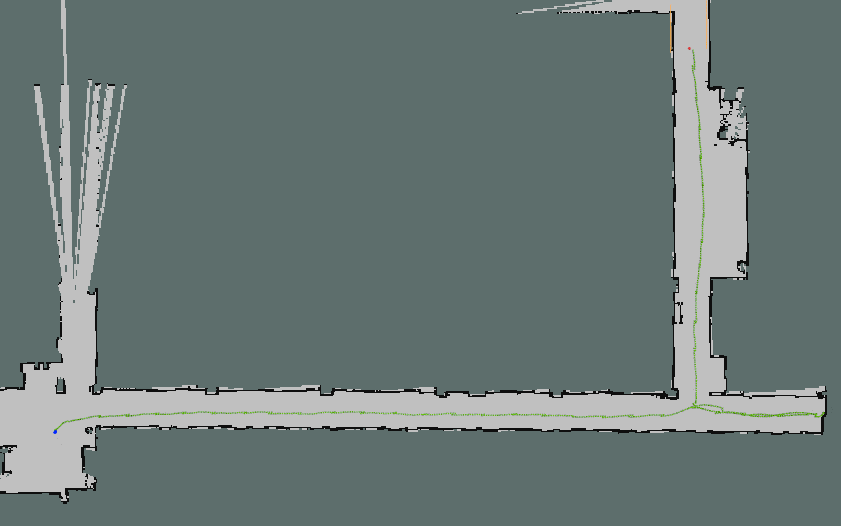}
    \caption{Sim-to-real results of the first experiment using our proposed navigation police for the robot navigating in the hallways of the UFMG School of Engineering (located in Belo Horizonte - MG, Brazil). The green line represents the robot's path from the starting (blue) point to the target (red) point. The grey area is the free space on the map, black points are obstacles, and dark green is the unknown area or places to be discovered.}
    \label{fig:expReal_1}
\end{figure}

\begin{figure}[t]
    \centering
    \includegraphics[clip,width=1\linewidth]{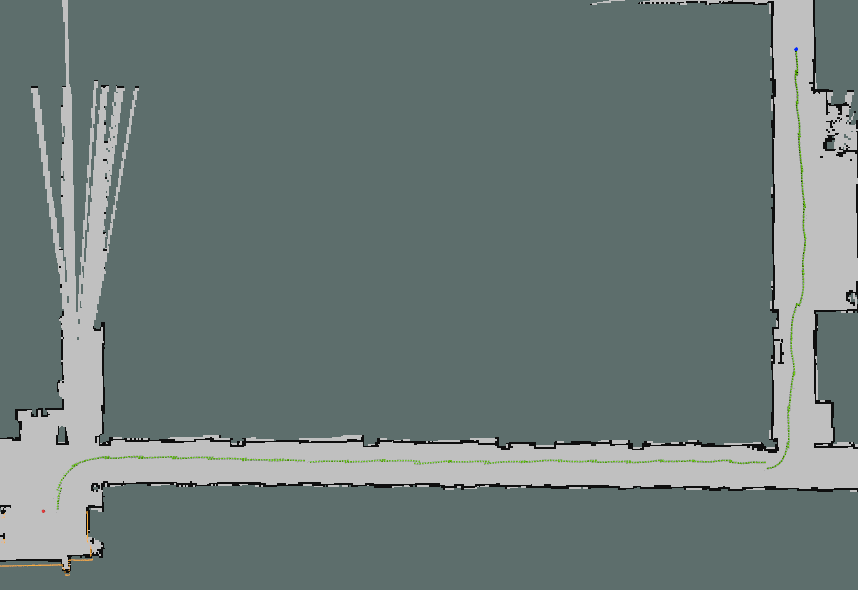}
    \caption{Sim-to-real results of the second experiment using our proposed navigation police for the robot navigating in the hallways of the UFMG School of Engineering. The green line represents the robot's path from the starting (blue) point to the target (red) point. The grey area is the free space on the map, black points are obstacles, and dark green is the unknown area or places to be discovered.}
    \label{fig:expReal_2}
\end{figure}

Fig.~\ref{fig:video_frame} presents the performed experiment in a sequence of frames.
Fig.~\ref{fig:expReal_1} and \ref{fig:expReal_2} show the path performed by the robot in two experiments illustrated by the green line, starting on the blue point and finishing on the red point. Additionally, a map was created simultaneously using the software Gmapping\footnote{Gmapping ROS - \url{http://wiki.ros.org/gmapping}} only for visual feedback of the environment. The results in Fig.~\ref{fig:expReal_1} show that, in the first moment, the trained policy missed the passage on the left for the target and passed through straight. However, the robot does not get stuck in a local minima, returning and arriving at the target. In the experiment in Fig.~\ref{fig:expReal_2}, the robot goes directly to the target, avoiding obstacles (black points on the map).

In the first experiment, the robot navigates a distance of $105.34~\mathrm{m}$, and in the second $87.47~\mathrm{m}$. Using the proposed local navigation police, the robot was capable of successfully navigating towards the goal region in unknown cluttered environments, avoiding collisions and local minima. The first distance traveled is greater due to the passage lost during the experiment. A video illustrating the training, sim-to-sim, and sim-to-real experiments can be seen in \protect\url{https://youtu.be/U2s61JZ_wTQ}.

% \protect\url{https://youtu.be/rI7XS0sqkqM}.

% The tests were performed using \ac{ROS} Melodic and Ubuntu 18.04, now running in a \emph{Jetson TX2} NVIDIA Pascal GPU architecture with 256 CUDA cores. The robot was a \emph{Pioneer 3at} equipped with a \emph{Hokuyo UTM-30LX-EW} laser scanner.

%

\section{Conclusion and future work}
\label{sec:conclusion}

In this paper, we have proposed a novel reward function for reinforcement learning and used the \acd{SAC} algorithm to train a \ac{DRL} policy in the context of local navigation for autonomous mobile robots in unknown cluttered environments. Our method improved the generalization of the policy, in the sense that it was capable of navigating the robot throughout the workspace with higher success rates than others existing approaches. By success, we mean the ability to reach the target region avoiding collisions and local minima.
Our first conclusion is that using the surrounding (\ac{LiDAR}) information in the reward function significantly improves the success rate when compared with others \cite{cimurs2021goal,hu2020voronoi,grando2021deep}. In the reward, the map information gained during navigation prevents actions that do not provide an exploration of new places, and the distance to the target prevents navigation away from it. 
Another conclusion is that the \ac{SAC} outperforms the \ac{TD3} for almost all reward functions used here and in the previous studies, justifying the use of learning algorithms that adopt techniques for increasing action exploration.

Regarding future works, we intend to study the use of asynchronous algorithms for learning by using the proposed reward function to optimize the process, including a high number of scenarios, and evaluate its influence on generalization.
Additionally, we want to perform experiments as an ablation study, checking the effects of noise in the sensor measures in the policy.

Autonomous exploration tasks could be carried out as future work, considering the presented navigation policy and another policy trained with the proposed strategy for frontier selection to increase the map information gain and reduce the exploration time.

\begin{acknowledgements}

This work has been financed by the Coordenação de Aperfeiçoamento de Pessoal de Nível Superior - Brasil (CAPES), Conselho Nacional de Desenvolvimento Científico e Tecnológico - Brasil (CNPq), and the Fundação de Amparo à Pesquisa do Estado de Minas Gerais (FAPEMIG). Also, we gratefully acknowledge the support of NVIDIA Corporation with the donation of the Jetson TX2 GPU used for this research.

\end{acknowledgements}

% \bibliographystyle{spmpsci}      % mathematics and physical sciences
% \bibliographystyle{spphys}       % APS-like style for physics
% \bibliographystyle{IEEEtran}
\bibliographystyle{elsarticle-num}

\bibliography{refs.bib}   % name your BibTeX data base
% \bibliography{refs.bbl}   % name your BibTeX data base

% \bibliography{main.bbl} 

\end{document}